\documentclass{article}

\usepackage[preprint]{neurips_2024}

\usepackage[utf8]{inputenc} % allow utf-8 input
\usepackage[T1]{fontenc}    % use 8-bit T1 fonts
\usepackage{hyperref}       % hyperlinks
\usepackage{url}
\usepackage{booktabs}       %
\usepackage{amsfonts}       %
\usepackage{nicefrac}       %
\usepackage{microtype}      %
\usepackage{amsmath,amsthm}
\usepackage{graphicx} % For including images
\usepackage{caption}
\usepackage{subcaption}

\usepackage{pifont}  
\usepackage{wrapfig}
\usepackage[font=small]{caption} 
\usepackage{listings}
\usepackage{algorithm}
\usepackage{algorithm}  
\usepackage{algorithmicx}
\usepackage{algpseudocode}  
\usepackage{subcaption}

\usepackage{xspace}
\usepackage{bbm}
\usepackage{xcolor}

\usepackage[toc,page,header]{appendix}
\usepackage{minitoc}
\usepackage{xpatch}

\usepackage{subcaption}
\usepackage{multirow, makecell}
\usepackage{multicol}
\usepackage{array}
\usepackage{amssymb,bm,amsthm}
\usepackage{enumitem}
\usepackage[misc,geometry]{ifsym}
\usepackage{adjustbox}
\usepackage{xfrac}
\usepackage{listings}
\usepackage{ltablex}  %
\usepackage{tabularx}

\usepackage[capitalize]{cleveref}
\crefname{section}{Sec.}{Secs.}
\Crefname{section}{Section}{Sections}
\Crefname{table}{Table}{Tables}
\crefname{table}{Tab.}{Tabs.}
\Crefname{algorithm}{Alg,}{Algs}
\crefname{algorithm}{Algorithm}{Algorithms}

\newcommand{\ourmethod}{RoboCoder\xspace}

\usepackage{listings}
\usepackage{multicol}
\usepackage{color}

\definecolor{codegreen}{rgb}{0,0.6,0}
\definecolor{codegray}{rgb}{0.5,0.5,0.5}
\definecolor{codepurple}{rgb}{0.58,0,0.82}
\definecolor{backcolour}{rgb}{0.95,0.95,0.92}

\usepackage{enumitem}
\setenumerate{itemsep=0pt,partopsep=0pt,parsep=0pt,topsep=0pt}

\newcommand{\pgchen}[1]{\textcolor{black}{#1}}

\lstdefinestyle{mystyle}{
    backgroundcolor=\color{backcolour},   
    commentstyle=\color{codegreen},
    keywordstyle=\color{magenta},
    numberstyle=\tiny\color{codegray},
    stringstyle=\color{codepurple},
    basicstyle=\footnotesize,
    breakatwhitespace=false,         
    breaklines=true,                 
    captionpos=b,                    
    keepspaces=true,                 
    numbers=left,                    
    numbersep=5pt,                  
    showspaces=false,                
    showstringspaces=false,
    showtabs=false,                  
    tabsize=2
}

\title{
\ourmethod: Robotic Learning from Basic Skills to General Tasks with Large Language Models}
\author{
 \textbf{Jingyao Li\textsuperscript{1}},
 \textbf{Pengguang Chen\textsuperscript{2}},
 \textbf{Sitong Wu\textsuperscript{1}},
 \textbf{Chuanyang Zheng\textsuperscript{1}},
 \textbf{Hong Xu\textsuperscript{1}},
 \textbf{Jiaya Jia\textsuperscript{1,2}}
\\
\\
  \textsuperscript{1}CUHK,
  \textsuperscript{2}SmartMore
}

\begin{document}

\maketitle

\begin{abstract}
% The emergence of Large Language Models (LLMs) has improved the prospects for robotic tasks. However, existing benchmarks are still limited to single tasks with limited generalization capabilities. In this work, we introduce a comprehensive benchmark and an autonomous learning framework, RoboCoder aimed at enhancing the generalization capabilities of robots in complex environments. Unlike traditional methods that focus on single-task learning, our research emphasizes the development of a general-purpose robotic coding algorithm that enables robots to leverage basic skills to tackle increasingly complex tasks. The newly proposed benchmark consists of 80 manually designed tasks across 7 distinct entities, testing the models' ability to learn from minimal initial mastery. Initial testing revealed that even advanced models like GPT-4 could only achieve a 47% pass rate in three-shot scenarios with humanoid entities. To address these limitations, the RoboCoder framework integrates Large Language Models (LLMs) with a dynamic learning system that uses real-time environmental feedback to continuously update and refine action codes. This adaptive method showed a remarkable improvement, achieving a 36% relative improvement. Our codes will be released.
The emergence of Large Language Models (LLMs) has improved the prospects for robotic tasks. However, existing benchmarks are still limited to single tasks with limited generalization capabilities. In this work, we introduce a comprehensive benchmark and an autonomous learning framework, \ourmethod, aimed at enhancing the generalization capabilities of robots in complex environments. Unlike traditional methods that focus on single-task learning, our research emphasizes the development of a general-purpose robotic coding algorithm that enables robots to leverage basic skills to tackle increasingly complex tasks. The newly proposed benchmark consists of 80 manually designed tasks across 7 distinct entities, testing the models' ability to learn from minimal initial mastery. Initial testing revealed that even advanced models like GPT-4 could only achieve a 47\% pass rate in three-shot scenarios with humanoid entities. To address these limitations, the \ourmethod framework integrates Large Language Models (LLMs) with a dynamic learning system that uses real-time environmental feedback to continuously update and refine action codes. This adaptive method showed a remarkable improvement, achieving a 36\% relative improvement. Our codes will be released.
\end{abstract}

%%%%%%%%%%%%%%%%%%%%%%%%%%%%%%%%%%%%%%%%%%%%%%%%%%%%%%%%%%%%
\section{Introduction}
%%%%%%%%%%%%%%%%%%%%%%%%%%%%%%%%%%%%%%%%%%%%%%%%%%%%%%%%%%%%

The emergence of Large Language Models (LLMs) has improved the prospects for robotic tasks. However, existing benchmarks are still limited to single tasks, with each task being independently trained. This results in limited generalization capabilities of current research~\cite{yu2023language, eureka}, only touching the surface of the inherent potential of these models.

In contrast, our research focuses on exploring general-purpose robotic coding algorithms. We are not only focused on excelling at individual tasks but also aim for the model to autonomously use existing skills to learn more complex tasks with limited initial mastery, thereby enhancing the model's applicability in real-world scenarios.

\begin{figure*}[t]
\centering 
\includegraphics[width=\textwidth, trim=0 0 0 0, clip]{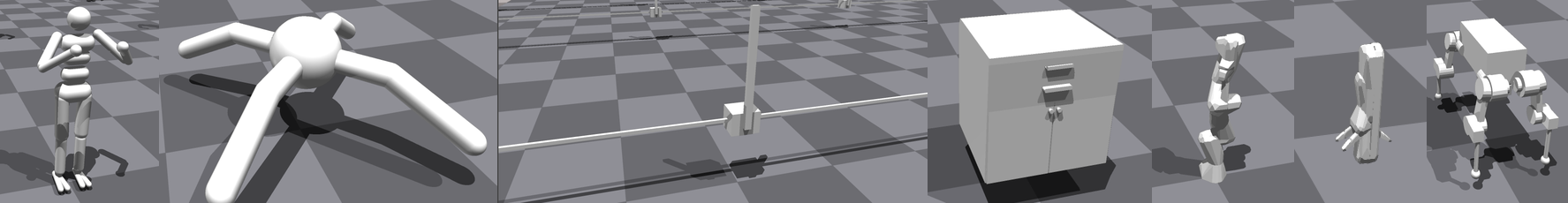}
\caption{Diverse Simulation Environments showcasing models used for testing robotic and biomechanical frameworks. (a) Human: a bipedal humanoid model; (b) Ant: a multi-legged robotic creature; (c) Cartpole: a classic control system with a pole balanced on a moving cart; (d) Sektion Cabinet: a static storage unit with drawable components; (e) Franka Panda: a robotic arm with articulation for intricate tasks; (f) Kinova: a modular robotic arm with advanced maneuverability; (g) Anymal: a quadruped robot designed for versatile mobility across varying terrain.}
\label{fig:envs}
\end{figure*}

Motivated by this, we propose a new benchmark that allows the model to leverage a few simple skills to first complete the writing of complex model codes in a few-shot manner, tested within the environment, and permits updates to the current action code based on environmental feedback within a limited budget. We uses the action images input from the simulated entities within the environment as a testing standard for the model, to verify if the model can accomplish the designated tasks. We have constructed a comprehensive benchmark consisting of 80 manually designed tasks across 7 distinct entities~\cite{isaac}. These range from humanoid figures and insect-like robots to complex mechanical arms and dynamic control systems, as depicted in \cref{fig:envs}. Testing with our benchmark, even complex coding models like GPT-4 only achieve a pass rate of 47\% in three-shot scenarios with humanoid entities. 

To better adapt to these more complex and realistic scenarios, we propose an adaptive learning framework, \ourmethod, which overcomes the previous bottlenecks encountered in learning new tasks, enabling effective utilization of environmental feedback for automatic updates. As illustrated in \cref{fig:framework}, RoboCoder consists of three basic components: \emph{Searcher}: When the target task is input into the searcher, it first retrieves the action space vector space for similarity to the target task. If the similarity exceeds a certain threshold, it is assumed that the target task exists within the action space, and the corresponding action code is output. \emph{Actor}: If the threshold is not exceeded, the searcher outputs the $k$ action codes exceeding the lower limit threshold to the Actor, who then refines and updates the input action codes to generate candidate actions that meet the target task. \emph{Evaluator}: Finally, the evaluator checks the candidate actions, and if passed, outputs the final action and updates it within the action space. If not passed, the evaluator provides a solution for the Actor to modify the candidate actions until they meet the evaluator's standards or reach the maximum number of iterations.

Compared with GPT-4, our RoboCoder framework achieves a relative improvement of 36\% with humanoid entities. In complex environments, such as those involving quadruped robot dogs, this rate soars to an impressive 92\%, highlighting our model's ability to maximize benefits from environments that leverage common sense and complexity. This demonstrates our model's potential applications in the open world, showcasing the powerful adaptability and advancement offered by our system.

In summary, our contributions are as follows:

\begin{enumerate}
     \item We have proposed a new comprehensive benchmark, containing 80 manually designed tasks across 7 different entities, verifying the model's ability to learn complex tasks from simple ones.
     \item We introduce \ourmethod, an autonomous robotic learning framework that combines LLMs, allowing robots to expand their action range through the use of real-time feedback, thus enhancing the model's applicability in open scenarios.
     \item Our results showcase the capability of this framework to surpass coding models like GPT-4 in few-shot scenarios, particularly in complex environments.
\end{enumerate}

%%%%%%%%%%%%%%%%%%%%%%%%%%%%%%%%%%%%%%%%%%%%%%%%%%%%%%%%%%%%
\section{Related Works}
%%%%%%%%%%%%%%%%%%%%%%%%%%%%%%%%%%%%%%%%%%%%%%%%%%%%%%%%%%%%
\paragraph{Models for Action Output.}
Previous research has predominantly explored the integration of vision and language inputs within embodied environments, aiming for the direct prediction of actions. Such studies~\cite{guhur2022instruction, shridhar2022perceiver, zhang2021hierarchical, silva2021lancon, jang2022bc, brohan2022rt} have made significant strides, with notable examples like VIMA~\citep{jiang2022vima} investigating the use of multimodal prompts and PaLM-E~\cite{palme} generating high-level textual instructions to harness the model's inherent world knowledge through self-conditioning on its predictions. However, these efforts predominantly revolve around high-level semantic planning for actions. Despite the progress in understanding and generating complex instructions, the direct conversion of these instructions into executable robotic actions remains an area ripe for exploration. This gap underscores a crucial challenge: bridging the divide between conceptual action planning and the practical implementation of these plans in real-world robotic tasks.

\paragraph{Models for Robot Code Generation.}
Recent research has explored the potential of coding Large Language Models (LLMs) to produce grounded and structured programmatic outputs that address decision-making and robotics challenges. These studies~\citep{austin2021program, chen2021evaluating, codellama, liang2023code, singh2023progprompt, wang2023demo2code, huang2023instruct2act, wang2023voyager, liu2023llm+, silver2023generalized, ding2023task, lin2023text2motion, xie2023translating, robocodex} have primarily utilized known motion primitives to execute robotic actions, rather than innovating and refining the actions autonomously. The most recent advancements~\citep{yu2023language, eureka} have investigated the support of LLMs in designing rewards for reinforcement learning scenarios, yet these approaches require extensive learning efforts and face limitations in task generalization. Our methodology stands out as the inaugural effort to harness LLMs for learning motion primitives, thereby establishing a dynamic action space. It circumvents the need for reinforcement learning, thereby freeing it from the constraints of previously learned environments.

\paragraph{Models for Evolution.}
The integration of evolutionary algorithms with LLMs has been a subject of investigation in various domains, including neural architecture search~\citep{chen2023evoprompting, nasir2023llmatic}, prompt engineering~\citep{guo2023connecting}, morphology design~\citep{lehman2022evolution}, and reward design~\cite{yu2023language, eureka}. \ourmethod innovates within this landscape by introducing an action space updating mechanism that facilitates the model's self-evolution and iteration, significantly enhancing the system's success rate and operational speed. It leverages the adaptability and learning capabilities of LLMs to continuously improve and refine actions.

\paragraph{Models for Code Repair.}
The use of large language models for program synthesis has been studied extensively in the literature \cite{alphacode, austin2021program, codex, codeRL, fried2022incoder, codegen, chowdhery2022palm, touvron2023llama, li2023starcoder}. There is also much contemporary work seeking to self-repair with LLMs \cite{zhang2023selfedit, hendrycksapps2021, codex, fried2022incoder, codegen, chen2023teaching, madaan2023selfrefine}. However, these methods are mostly aimed at simple programming tasks, and there has been no discussion on how to effectively translate them into machine language. In contrast, our method for the first time constructs an action space, which can significantly improve the model's inference speed and accuracy with an increase in the number of iterations, and is easily extendable to new tasks and entities.

%%%%%%%%%%%%%%%%%%%%%%%%%%%%%%%%%%%%%%%%%%%%%%%%%%%%%%%%%%%%
\section{Methods}
%%%%%%%%%%%%%%%%%%%%%%%%%%%%%%%%%%%%%%%%%%%%%%%%%%%%%%%%%%%%
\begin{figure*}[t]
\centering 
\includegraphics[width=\textwidth, trim=90 120 100 170, clip]{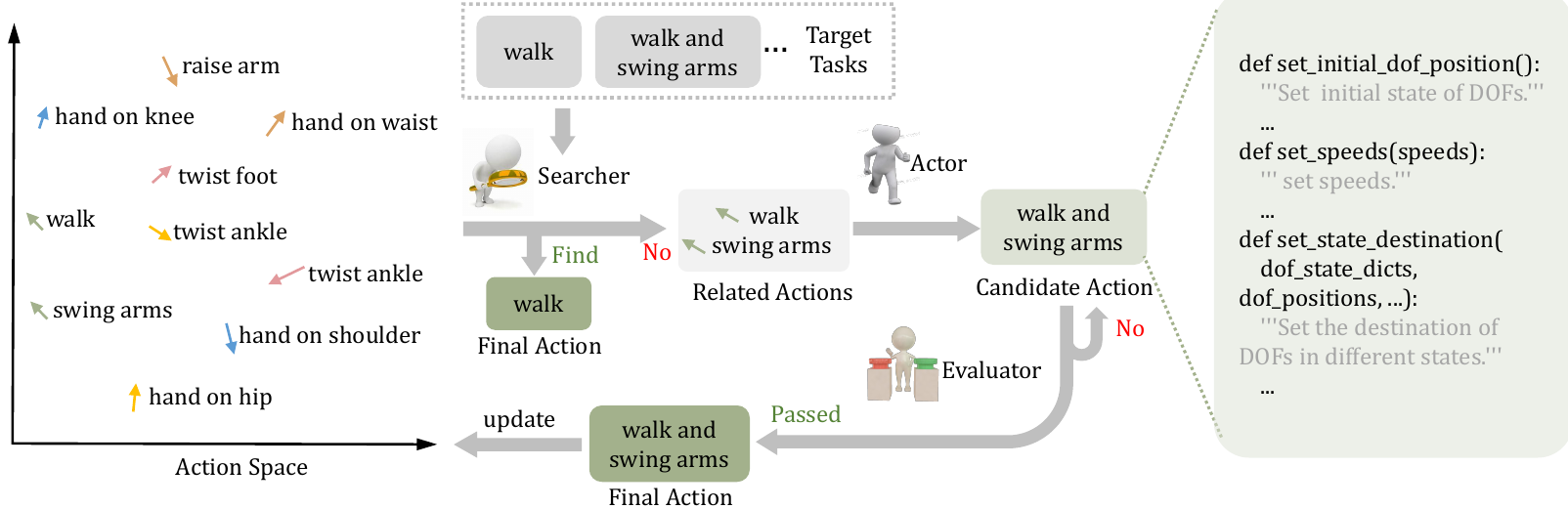}
\caption{\textbf{Searcher}: When a target task is input to the searcher, the searcher first retrieves the similarity within the action space's vector space to the target task. If the similarity exceeds the upper threshold, it is considered that the target task exists within the action space, and the corresponding action code is output. \textbf{Actor}: If the upper threshold is not exceeded, the searcher outputs $k$ action codes that surpass the lower threshold to the actor, who then improves and updates the input action codes to generate candidate actions that satisfy the target task. \textbf{Evaluator}: Finally, the evaluator examines the candidate actions, and if passed, the final action is output and updated in the action space. If not passed, the evaluator provides a solution for the actor to revise the candidate action until it meets the evaluator's standards or reaches the maximum number of iterations.}
\label{fig:framework}
\end{figure*}

\iffalse
The \ourmethod framework, as illustrated in \cref{fig:framework}, comprises three key components: 
\begin{enumerate}[leftmargin=*,labelsep=8pt]
    \item \textbf{Action Searcher}: A lightweight language model designed to assess the similarity between the target task and the action space, facilitating the retrieval of similar actions. 
    \item \textbf{Actor}: A large language model responsible for refining and updating related actions to produce candidate actions that meet the requirements of the target task. 
    \item \textbf{Evaluator}: a vision language model tasked with evaluating the candidate actions. 
\end{enumerate}
\fi 

The \ourmethod framework is illustrated in \cref{fig:framework}. \pgchen{We have defined an initial action space, which includes some fundamental actions. Through our carefully designed process, our framework builds upon this initial action space, gradually inferring more complex target actions.} Initially, the searcher evaluates the given target action against the action space to find similar actions; if a direct match is found, the appropriate action code is promptly outputted. In scenarios where the similarity does not surpass this threshold, the Searcher then identifies related actions, which the Actor refines to formulate candidate actions that align with the task's requirements. Finally, the Evaluator scrutinizes these candidate actions for compliance with the set criteria. Successful actions are integrated into the action space, whereas those falling short are sent back for revision by the Actor.

%%%%%%%%%%%%%%%%%%%%%%%%%%%%%%%%%%%%%%%%%%%%%%%%%%%%%%%%%%%%
\subsection{Action Search}
\label{sec:searcher}
%%%%%%%%%%%%%%%%%%%%%%%%%%%%%%%%%%%%%%%%%%%%%%%%%%%%%%%%%%%%
In this section, we introduce the action search process. We first define an initial action space, as illustrated in \cref{sec:action_space}. We detail only some actions of Humans, while for other entities, we simply provide a few examples in a few-shot manner to verify the model's self-expansion and update capabilities.

The Action Searcher receives the target task with the aim of retrieving actions from the action space that are either identical to or closely related to the target task, and providing them to the actor. It employs the Sentence Transformer to encode both the target task and the action examples into a shared vector space. The core idea revolves around measuring the cosine similarity between the target task embedding and each action example embedding, thus identifying the most relevant actions.

First, given a target task $t$ and an action space $\mathbf{A} = \{\mathbf{a}_1, \mathbf{a}_2, \ldots, \mathbf{a}_n\}$, where $n$ is the number of actions in the action space, we compute the embedding of the target task, $\mathbf{e}_t$, and the embeddings of each action in the action space $\mathbf{A}$, $\mathbf{E}_a = \{\mathbf{e}_{a_1}, \mathbf{e}_{a_2}, \ldots, \mathbf{e}_{a_n}\}$, using the encoder model $f_e$:
\begin{equation}
    \mathbf{e}_t = f_e(t), \mathbf{e}_{a_i} = f_e(\mathbf{a}_i),
\end{equation}

Then the cosine similarity $\mathbf{c} = \{c_{1}, c_{2}, \ldots, c_{n}\}$ between $\mathbf{e}_t$ and each $\mathbf{e}_{a_i}$ is calculated as follows:
\begin{equation}
    c_i = \frac{\mathbf{e}_t \cdot \mathbf{e}_{a_i}}{\|\mathbf{e}_t\| \|\mathbf{e}_{a_i}\|}.
\end{equation}

After obtaining the cosine similarity, we proceed to select the action that is either identical to or closely related to the target task, and provide them to the actor. We propose a dual-threshold selection strategy:
\begin{enumerate}
    \item[Case 1.] If there are exceptionally high scores, for example, $\exists c_i > \Lambda_h$ \pgchen{(which is set to 0.99 in practice)}, it is considered that the corresponding target task's action has been retrieved in the action space. Thus, the actor will directly output the corresponding action $a_i$, without the need to regenerate the action through the LLM.
    \item[Case 2.] If there are relatively high scores, for example, $\exists c_i > \Lambda_l$\pgchen{ (which is set to 0.5 in practice)}, it is considered that actions related to the target task exist in the action space. At this point, the top $k$ actions with the highest scores will be output. If there are fewer than $k$ actions, all actions with $c_i > \Lambda_l$ will be output.
    \item[Case 3.] If $\forall c_i < \Lambda_l$, it is considered that all actions in the action space are irrelevant to the target task. In this case, only the closest action will be output as a template to guide the LLM in generating the action.
\end{enumerate}

Formally, the selection process can be formalized as:
\begin{equation}
\mathbf{A}_f = \begin{cases} 
   \{\mathbf{a}_i\} & \text{if } \exists c_i > \Lambda_h, \\
   \{\mathbf{a}_i \ | \ c_i \in \text{topk}(\mathbf{c}, k) \}  \setminus \{\mathbf{a}_i \ | \ c_i < \Lambda_l\}  & \text{elif } \exists c_i > \Lambda_l, \\
   \{\mathbf{a}_i \ | \ c_i \in \text{topk}(\mathbf{c}, 1) \} & \text{otherwise,}
\end{cases}
\end{equation}
where $\text{topk}(\mathbf{c}, k)$ retrieves the top $k$ scores from $\mathbf{c}$, and $\mathbf{A}_f$ represents the final set of selected actions. If any action $\mathbf{a}_i$ in the top $k$ actions has a similarity score $c_i$ greater than the upper threshold $\Lambda_h$, $\mathbf{a}_i$ is immediately considered as the final action. If no action's score surpasses $\Lambda_h$, the set $\mathbf{A}_k$ is filtered to remove any actions with scores below the lower threshold $\Lambda_l$, in which case all actions above $\Lambda_l$ are considered. If none of the actions' scores exceed \(\Lambda_l\), then output the action with the highest score. The action selector algorithm is delineated in \cref{sec:alg}.

\begin{figure*}[t]
\centering 
\includegraphics[width=\textwidth, trim=70 130 280 130, clip]{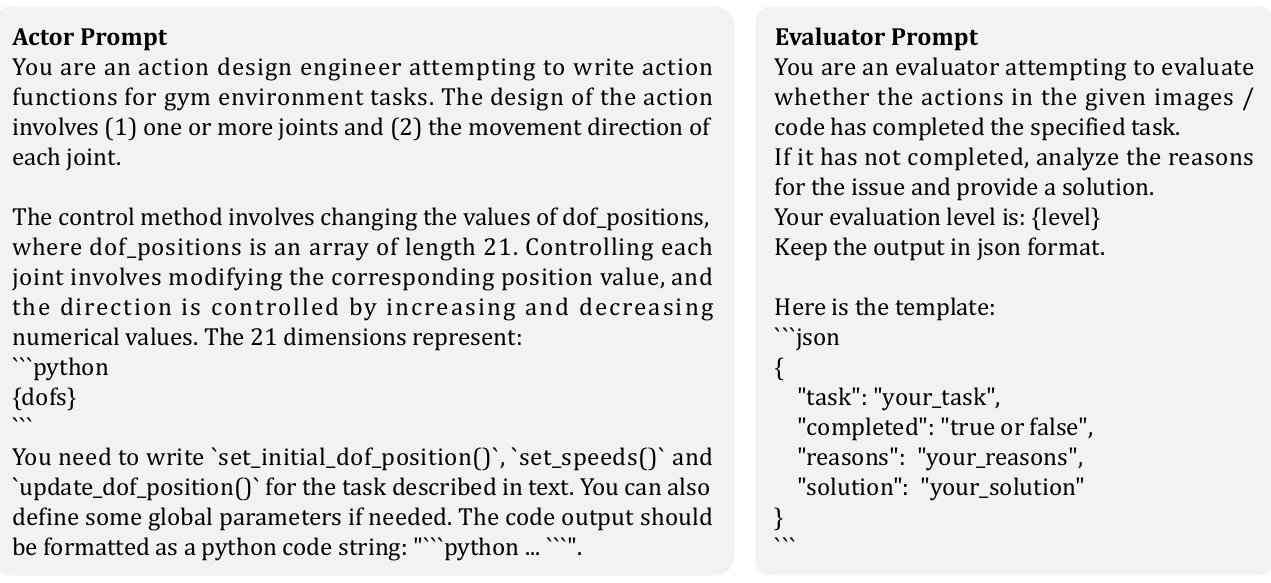}
\caption{The actor and evaluator system prompts.}
\label{fig:prompts}
\end{figure*}

\begin{figure*}[t]
\centering 
\includegraphics[width=\textwidth, trim=70 130 240 130, clip]{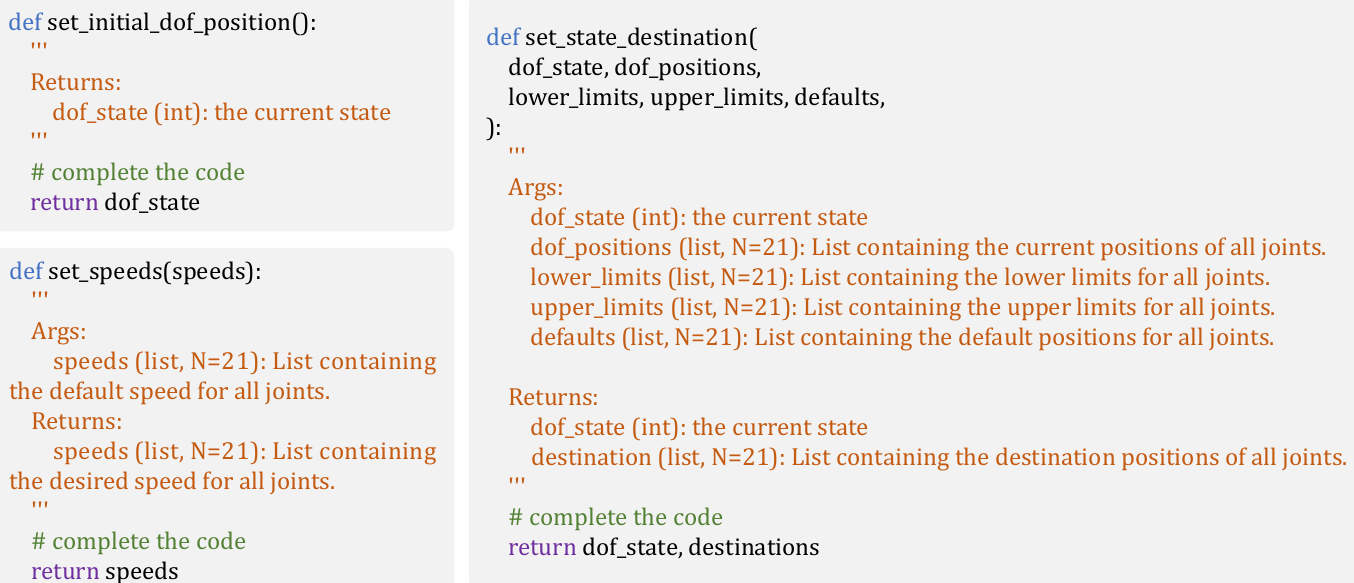}
\caption{The templates of the action code.}
\label{fig:template}
\end{figure*}

%%%%%%%%%%%%%%%%%%%%%%%%%%%%%%%%%%%%%%%%%%%%%%%%%%%%%%%%%%%%
\subsection{Actor}
\label{sec:actor}
%%%%%%%%%%%%%%%%%%%%%%%%%%%%%%%%%%%%%%%%%%%%%%%%%%%%%%%%%%%%
\pgchen{The Actor, powered by a Large Language Model (LLM), aims to generate target actions based on similar actions selected within the Searcher.}

The input to the Actor includes prompts containing the actions from the action searcher, descriptions of degrees of freedom (DOFs), and information about the target task. The DOFs are detailedly introduced in \cref{sec:dofs}, providing the Actor with the type of motion for each DOF, limit positions, and direction of movement. The Actor's system prompt is depicted in \cref{fig:prompts}.

\pgchen{As required by the prompt,} the Actor will output the action function, including three essential functions, each illustrated in the template provided in \cref{fig:template}. The function \textit{set\_initial\_dof\_position} sets the initial state for each joint. The function \textit{set\_speeds} regulates the speed at which each joint transitions between positions. The function \textit{set\_state\_destination} delineates the target positions for each DOF and orchestrates the necessary state transitions to reach the predetermined end state. Together, these functions ensure comprehensive control over the movements of the simulated entity.

%%%%%%%%%%%%%%%%%%%%%%%%%%%%%%%%%%%%%%%%%%%%%%%%%%%%%%%%%%%%
\subsection{Evaluator}
\label{sec:evaluator}
%%%%%%%%%%%%%%%%%%%%%%%%%%%%%%%%%%%%%%%%%%%%%%%%%%%%%%%%%%%%

\begin{figure*}[t]
\centering 
\includegraphics[width=\textwidth, trim=60 140 135 200, clip]{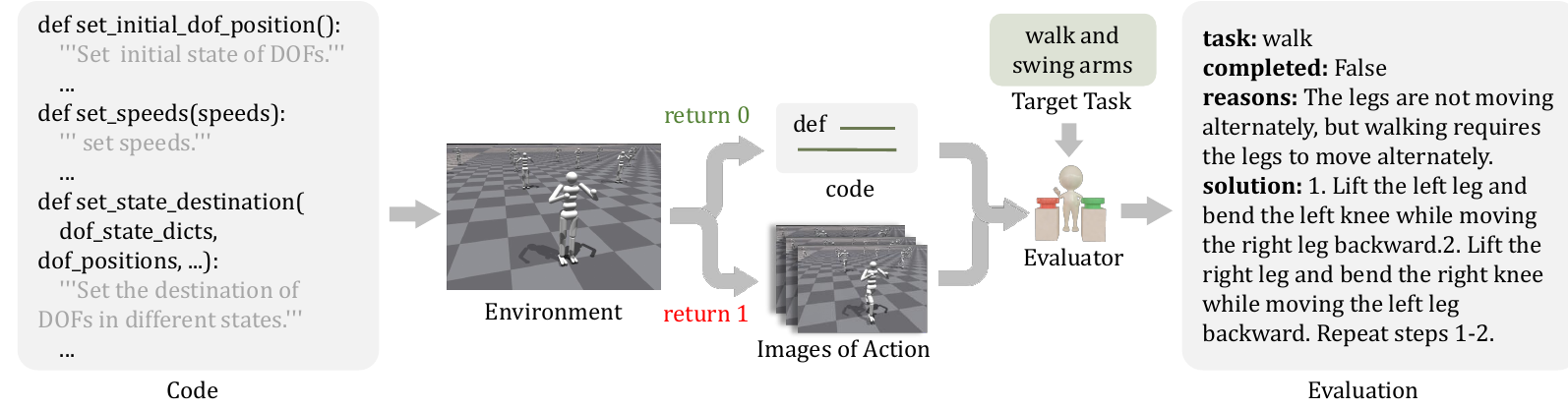}
\caption{\textbf{Process of evaluation.} When an action code is tested in the target environment, if the environment returns 1, the system outputs the code along with error messages. If the environment returns 0, it output images of the action operation. The evaluator receives the information and determines whether the target task has been successfully completed. If the target task is not successfully completed, a solution is provided to the actor for improvement.}
\label{fig:evaluator}
\end{figure*}

In this section, we introduce the Evaluator, which leverages a vision-language model to determine whether the  Actor's output action completes the specified task, as depicted in \cref{fig:evaluator}. We evaluate 7 distinct entities in the IsaacGym~\cite{isaac}, from humanoid figures and insect-like robots to complex mechanical arms and dynamic control systems, as depicted in \cref{fig:envs}. When actions are executed in the environment, the environment's response is binary:
\begin{enumerate}
\item[Return 0] indicates a successful execution of the action. In this case, the Evaluator determines whether the entity has completed the target task based on images taken during the entity's action execution. If not completed, suggestions are made based on the entity's \pgchen{output}.
\item[Return 1] indicates the presence of errors. In this scenario, images cannot be captured; hence, the Evaluator analyzes the situation based on the action code and error messages, providing feedback and improvement strategies for the actor.
\end{enumerate}
In both scenarios, the evaluator receives images or codes of action, and outputs a json containing information such as \textit{completed}, \textit{reasons}, \textit{solution}, etc. The evaluator's prompt is depicted in \cref{fig:prompts}. If the task is completed, the cycle ends; if not, suggestions for improvement are provided to guide the actor in updating the action. We used the following prompt: \textit{Your current action code does not fulfill the task: <task>. Here is the reason: <reason>. Here is the suggested solution: <solution>. Please rewrite the action functions.} The Evaluator plays a crucial role in the iterative updates within the \ourmethod framework, ensuring a continual enhancement of action codes until the target task is accomplished. 

%%%%%%%%%%%%%%%%%%%%%%%%%%%%%%%%%%%%%%%%%%%%%%%%%%%%%%%%%%%%
\subsection{Update Action Space}
\label{sec:update}
%%%%%%%%%%%%%%%%%%%%%%%%%%%%%%%%%%%%%%%%%%%%%%%%%%%%%%%%%%%%

During an iteration, the Searcher initially identifies similar actions to the target action within the action space. Subsequently, the Actor refines these candidates, which are then rigorously evaluated by the Evaluator. Actions that meet the Evaluator's criteria are subsequently incorporated into the action space:
\begin{equation}
    \mathbf{A}^{(i+1)} = \mathbf{A}^{(i)} \cup \mathbf{a}_p
\end{equation}
where $\mathbf{a}_p$ represents the action that passed the Evaluator's test, and $\mathbf{A}^{(i)}$ is the action space of the $i$-th iteration. 

A strategy for updating the action space is serial updating (\cref{tab:serial}), which involves updating the action space after each task is completed or when the maximum number of iterations is reached, before proceeding to the next task. In contrast, we employ a parallel updating strategy (\cref{tab:parallel}), where all tasks are tested simultaneously. Simple tasks can be quickly integrated into the action space, while difficult tasks can benefit from insights gained from simpler ones. After each iteration, we update the action space before moving forward.

As the number of iterations increases, leading to a more extensive action space, we are able to leverage a lightweight action searcher to promptly match verified code within the action space. This negates the need for rewriting code via a large language model, thereby accelerating our inference process and enhancing inference accuracy. We will quantitatively demonstrate this mechanism through experimental evidence in \cref{sec:ablation}.

%%%%%%%%%%%%%%%%%%%%%%%%%%%%%%%%%%%%%%%%%%%%%%%%%%%%%%%%%%%%
\section{Experiments}
%%%%%%%%%%%%%%%%%%%%%%%%%%%%%%%%%%%%%%%%%%%%%%%%%%%%%%%%%%%%

%%%%%%%%%%%%%%%%%%%%%%%%%%%%%%%%%%%%%%%%%%%%%%%%%%%
% Ablation 1: reason feedback ablation
% Ablation 2: image evaluator ablation
%%%%%%%%%%%%%%%%%%%%%%%%%%%%%%%%%%%%%%%%%%%%%%%%%%%
\begin{figure}[t]
    \centering
    \includegraphics[width=\textwidth, trim=0 0 0 0, clip]{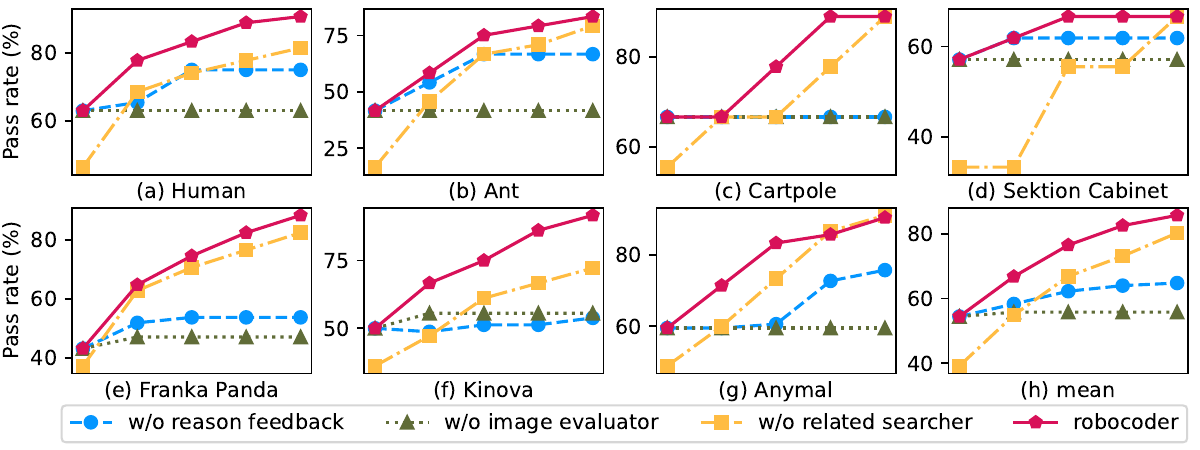}
    \captionof{figure}{Ablation with the Actor GPT4. \textit{w/o feedback} regenerates the action without reason feedback. \textit{w/o image feedback} leverages the action code and terminal information without the action images for feedback generation. \textit{w/o related searcher} utilizes the random searcher to select actions.}
    \label{fig:feedback}
\end{figure}

%%%%%%%%%%%%%%%%%%%%%%%%%%%%%%%%%%%%%%%%%%%%%%%%%%%
% Ablation 3: action space updates
% Performance 1:  More models performance
%%%%%%%%%%%%%%%%%%%%%%%%%%%%%%%%%%%%%%%%%%%%%%%%%%%
\begin{figure}[t]
    \centering
    \begin{minipage}[t]{0.26\textwidth}
        \centering
        \includegraphics[width=\textwidth, trim=0 0 0 0, clip]{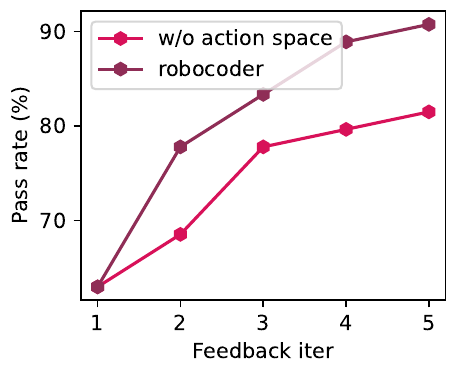}
        \captionof{figure}{Action space update ablation in cycle two with the Actor GPT3.5 for Human entity.}
        \label{fig:action-space}
    \end{minipage}
    \hfill\hfill
    \begin{minipage}[t]{0.26\textwidth}
        \centering
        \includegraphics[width=\textwidth, trim=0 0 0 0, clip]{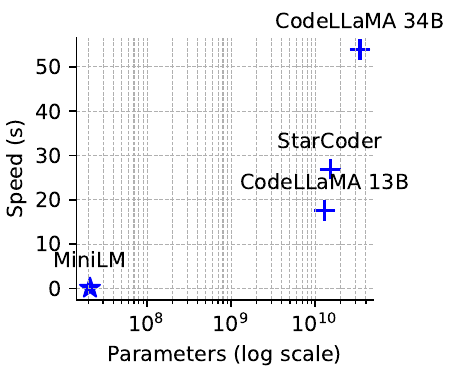}
        \captionof{figure}{Parameters and speed on a single A800 GPU of the Searcher and Actors for Human entity.}.
        \label{fig:speed}
    \end{minipage}
    \hfill\hfill
    \begin{minipage}[t]{0.43\textwidth}
        \centering
        \includegraphics[width=\textwidth, trim=0 0 0 0, clip]{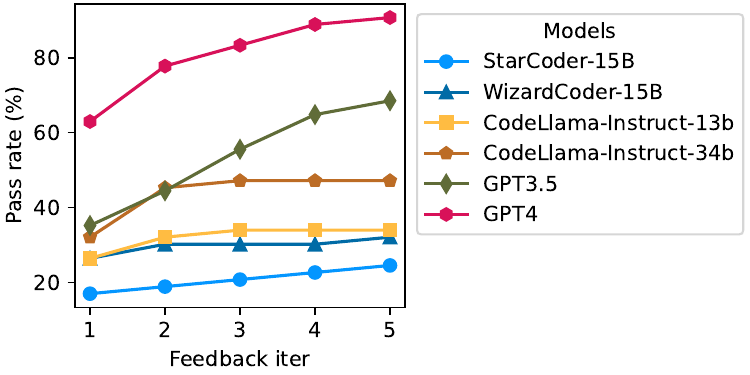}
        \captionof{figure}{Performance across models, including StarCoder-15B, WizardCoder-15B, CodeLLaMA-Instruct-13B and 34B, GPT3.5 and GPT-4 for Human entity.} % , including StarCoder-15B~\cite{starcoder}, WizardCoder-15B~\cite{wizardcoder}, CodeLLaMA-Instruct 13B and 34B~\cite{codellama}, GPT3.5\cite{gpt3.5} and GPT-4~\cite{openai2023gpt4}
        \label{fig:more-models}
    \end{minipage}
\end{figure}

\begin{figure}[t]
    \centering
    \includegraphics[width=\textwidth, trim=0 0 0 0, clip]{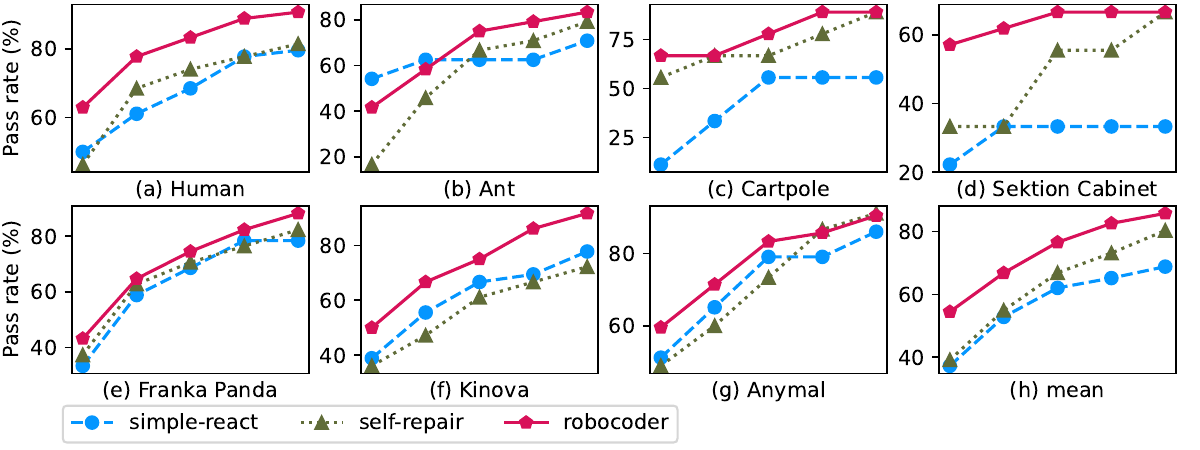}
    \captionof{figure}{Performance comparing with baseline and self-repair~\cite{shinn2023reflexion}.}.
    \label{fig:envs_performance}
\end{figure}

In this section, we first introduce the employment details of our experiment setting in \cref{sec:employment}. Then, we present our designed test benchmark to assess the framework's capability in handling a broad spectrum of movement in \cref{sec:benchmark}. Then, we undertake an ablation study, as detailed in \cref{sec:ablation}, to illuminate the role of each part of the framework. Furthermore, we evaluate the efficacy of our framework in \cref{sec:performance}, providing a comprehensive analysis of its performance enhancements and adaptability across various coding models. 

%%%%%%%%%%%%%%%%%%%%%%%%%%%%%%%%%%%%%%%%%%%%%%%%%%%%%%%%%%%%
\subsection{Employment Details}
\label{sec:employment}
%%%%%%%%%%%%%%%%%%%%%%%%%%%%%%%%%%%%%%%%%%%%%%%%%%%%%%%%%%%%

Our \ourmethod framework integrates the following specific models:

\textbf{Action Searcher}. We utilize MiniLM~\cite{minilm}, a lightweight language model. MiniLM stands out as a sentence-transformers model, adept at encoding sentences into a 384-dimensional dense vector space, enabling precise and scalable similarity assessments.

\textbf{Actor}. Central to our action generation process are SOTA coding LLMs, including StarCoder-15B~\cite{starcoder}, WizardCoder-15B~\cite{wizardcoder}, and CodeLLaMA-Instruct variants (13B and 15B)~\cite{codellama}, complemented by the advanced capabilities of the GPT series models, specifically GPT3.5\cite{gpt3.5} and GPT-4~\cite{openai2023gpt4}. We leverage GPT-4 for our default configuration as well as competitive methods. The initial action space is detailed in \cref{sec:action_space}.

\textbf{Evaluator}. To assess the effectiveness of generated actions, we employ GPT4V~\cite{openai2023gpt4}, a leading vision language model, ensuring evaluations are both comprehensive and convincing. It also offer actionable insights for the Actor to refine its action generation process. We conduct ablation experiments with GPT4~\cite{openai2023gpt4} in \cref{sec:code_evaluator}
.
%%%%%%%%%%%%%%%%%%%%%%%%%%%%%%%%%%%%%%%%%%%%%%%%%%%%%%%%%%%%
\subsection{Test Benchamrk}
\label{sec:benchmark}
%%%%%%%%%%%%%%%%%%%%%%%%%%%%%%%%%%%%%%%%%%%%%%%%%%%%%%%%%%%%
To evaluate the performance of our proposed RoboCoder, we introduce a test space consisting of 80 carefully designed tasks across diverse entities~\cite{isaac}, namely \textit{Human, Ant, Cartpole, Sektion Cabinet, Franka Panda, Kinova, and Anymal}. For example, the tasks of the \textit{Human} entity range from basic movements such as raising an arm to more complex activities like running. The detailed test tasks are in \cref{tab:test_space}. These tasks are intended to probe the baseline framework's ability to accurately simulate and understand the complexities of entities. By establishing a set of standardized tasks, we aim to encourage and facilitate subsequent studies to build upon and innovate beyond our baseline framework, pushing the boundaries of what is currently possible in the field.

%%%%%%%%%%%%%%%% human 
% Our environments consist of 18 tasks implemented using the Isaac Gym simulator~\citep{isaac}, encompassing a series of 18 meticulously designed target tasks aimed at evaluating the robustness and adaptability of our model. Each of the 18 sub-figures corresponds to a distinct task, ranging from executing rotations (\textit{right elbow, right knee, left ankle, left hip, right hip}) and simple gestures like raising a single arm or both arms, to more complex movements such as positioning the left hand on various parts of the body (\textit{left hip, left shoulder, left waist, right knee}),  and extending dynamic actions (\textit{stretching arms forward, swinging arms, and running}). These tasks were curated to cover a wide spectrum of movements, challenging the model's ability to accurately interpret and execute actions across different joints and movement patterns. This diverse set of tasks serves as a test benchmark for assessing the model's performance in simulating realistic and varied physical activities within a controlled environment.

%%%%%%%%%%%%%%%%%%%%%%%%%%%%%%%%%%%%%%%%%%%%%%%%%%%%%%%%%%%%
\subsection{Ablation Study}
\label{sec:ablation}
%%%%%%%%%%%%%%%%%%%%%%%%%%%%%%%%%%%%%%%%%%%%%%%%%%%%%%%%%%%%

\paragraph{Feedback.} We conducted ablation experiments on the feedback provided by the Evaluator. To demonstrate the benefit of the feedback provided for the Actor to update their action code, we conducted ablation experiments \textit{without feedback} (w/o feedback), which regenerates the action without any reason feedback. It uses the following simplified prompt: \textit{Your current action code does not fulfill the task: <task>. Please rewrite the action functions.} The results in \cref{fig:feedback} illustrate the benefits of reason feedback, with \ourmethod showing an average improvement of 20.82\% over \textit{w/o feedback} on entities.

\paragraph{Image Evaluator.} We conducted ablation experiments on the Evaluator. For \ourmethod, we utilize the Image Evaluator, which analyzes the behavior in images to provide feedback. We also conducted ablation with the Code Evaluator, commonly used in previous methods~\cite{shinn2023reflexion, zhang2023selfedit}, that uses action codes and terminal information without the action images for feedback generation. The results found that the Code Evaluator tends to mistakenly believe that the code has already completed the task, leading to stopping the iteration process at the first or second iteration, which results in the model's pass rate stagnating, with only a 1.35\% improvement compared to the first iteration. In contrast, after iterations, the RoboCoder's results improved by 31.27\%.  In \cref{sec:code_evaluator}, we provide the pass rate determined by the Code Evaluator.

\paragraph{Related Searcher.} We conducted ablation experiments on the Action Searcher. For our method, we employ the Related Searcher, which finds the most relevant action in the action space to provide to the Actor. To demonstrate the superiority of the Related Searcher, we also conducted ablation experiments on the Random Searcher, which selects actions randomly from the action space. The results in \cref{fig:feedback} prove the superiority of the Related Searcher over the Random Searcher, especially in the initial stages. It improved by 15.28\% in the first iteration and by 5.45\% at the end of the iterations.

\paragraph{Action Space Update.} The impact of action space updates on the performance of coding models was meticulously examined through an ablation study. Action space updates at the end of each iteration. The introduction of action space updates markedly enhanced performance, as shown in \cref{fig:action-space}. The results highlight the advantage of expanding the action space informed by previous actions and outcomes.

\paragraph{Parameters and Speed.} This two-stage analysis clarifies the significant benefits derived from leveraging learned actions to refine and expand the action space. The enlargement of the action space significantly bolsters both the pass rate and speed during inference. When a target task already exists within the action space, a swift match by a lightweight action searcher suffices for application. We test the speed of the searcher and actors. The results are presented in \cref{fig:speed}. a lightweight Searcher 21M MiniLMv2~\cite{minilmv2} can retrieve an action in merely 0.06 seconds, whereas a large language model like a 34B CodeLLaMA~\cite{codellama} requires up to 54 seconds. Therefore, the integration of an action searcher can dramatically accelerate inference speeds. This investigation reveals the superiority of our framework.

%%%%%%%%%%%%%%%%%%%%%%%%%%%%%%%%%%%%%%%%%%%%%%%%%%%%%%%%%%%%
\subsection{Performance}
\label{sec:performance}
%%%%%%%%%%%%%%%%%%%%%%%%%%%%%%%%%%%%%%%%%%%%%%%%%%%%%%%%%%%%

\paragraph{Models.} In our study, we observed remarkable performance enhancements across a spectrum of coding models, as delineated in the comparison of baseline and RoboCoder. The results, depicted in~\cref{fig:more-models}, showcases the progress made by integrating RoboCoder into various models, such as StarCoder-15B~\cite{starcoder}, WizardCoder-15B~\cite{wizardcoder}, CodeLLaMA-Instruct-13B and 34B~\cite{codellama}, along with the GPT series models~\cite{gpt3.5, openai2023gpt4}.

GPT4 series, in particular, demonstrated the most significant enhancement, registering a remarkable 30.56\% increase over its baseline performance. This is attributed to the more advanced models' superior capability to utilize environmental feedback for self-improvement. Such advancements not only highlight RoboCoder's effectiveness in boosting model performance but also underscore the potential for significant improvements in more sophisticated models. Across the board, a consistent and notable uplift in performance validates RoboCoder's robustness and adaptability in addressing a broad range of coding challenges.

\paragraph{Entities.} In our investigation, we documented significant and widespread improvements in our \ourmethod across various entities, extending from humanoid figures and insect-like robots to intricate mechanical arms and dynamic control systems. The results depicted in \cref{fig:envs_performance} indicate an average enhancement of 10.34\% compared to previous state-of-the-art methods. These findings affirm the potential of RoboCoder in adapting to and enhancing a wide array of robotic entities within diverse simulation environments.

\section{Limitation} 
%%%%%%%%%%%%%%%%%%%%%%%%%%%%%%%%%%%%%%%%%%%%%%%%%%%%%%%%%%%%
Due to hardware limitations and challenges in the physical world, our method is currently only implemented in simplified simulated environments and has not yet been validated in the real world. The tasks involved are also relatively simple and do not require extensive knowledge of the physical world. We hope our work can inspire future efforts dedicated to applying natural language to machine language for real-world robots.

%%%%%%%%%%%%%%%%%%%%%%%%%%%%%%%%%%%%%%%%%%%%%%%%%%%%%%%%%%%%
\section{Conclusion}
%%%%%%%%%%%%%%%%%%%%%%%%%%%%%%%%%%%%%%%%%%%%%%%%%%%%%%%%%%%%

In conclusion, we believe our work paves the path for addressing the limitations of existing benchmarks for robotic tasks by introducing a new comprehensive benchmark and the \ourmethod framework. Our contributions are twofold: first, the development of a benchmark consisting of diverse tasks across different entities, which rigorously tests the ability of robots to learn from minimal initial mastery and handle increasingly complex scenarios. Second, the implementation of the \ourmethod framework, which incorporates Large Language Models and an adaptive learning system, allows for dynamic updating of action codes based on real-time environmental feedback. This approach has demonstrated substantial improvements in performance, achieving a 36\% relative improvement in pass rates with humanoid entities and up to 92\% with more complex entities like quadruped robot dogs. These results not only validate the effectiveness of our methods but also highlight the potential for significant advancements in the applicability of robotic systems in real-world, open environments.

\bibliography{egbib}

\begin{thebibliography}{48}
\providecommand{\natexlab}[1]{#1}
\providecommand{\url}[1]{\texttt{#1}}
\expandafter\ifx\csname urlstyle\endcsname\relax
  \providecommand{\doi}[1]{doi: #1}\else
  \providecommand{\doi}{doi: \begingroup \urlstyle{rm}\Url}\fi

\bibitem[Austin et~al.(2021)Austin, Odena, Nye, Bosma, Michalewski, Dohan, Jiang, Cai, Terry, Le, and Sutton]{austin2021program}
Jacob Austin, Augustus Odena, Maxwell Nye, Maarten Bosma, Henryk Michalewski, David Dohan, Ellen Jiang, Carrie Cai, Michael Terry, Quoc Le, and Charles Sutton.
\newblock Program synthesis with large language models, 2021.

\bibitem[Brohan et~al.(2023)Brohan, Brown, Carbajal, Chebotar, Dabis, Finn, Gopalakrishnan, Hausman, Herzog, Hsu, Ibarz, Ichter, Irpan, Jackson, Jesmonth, Joshi, Julian, Kalashnikov, Kuang, Leal, Lee, Levine, Lu, Malla, Manjunath, Mordatch, Nachum, Parada, Peralta, Perez, Pertsch, Quiambao, Rao, Ryoo, Salazar, Sanketi, Sayed, Singh, Sontakke, Stone, Tan, Tran, Vanhoucke, Vega, Vuong, Xia, Xiao, Xu, Xu, Yu, and Zitkovich]{brohan2022rt}
Anthony Brohan, Noah Brown, Justice Carbajal, Yevgen Chebotar, Joseph Dabis, Chelsea Finn, Keerthana Gopalakrishnan, Karol Hausman, Alex Herzog, Jasmine Hsu, Julian Ibarz, Brian Ichter, Alex Irpan, Tomas Jackson, Sally Jesmonth, Nikhil~J Joshi, Ryan Julian, Dmitry Kalashnikov, Yuheng Kuang, Isabel Leal, Kuang-Huei Lee, Sergey Levine, Yao Lu, Utsav Malla, Deeksha Manjunath, Igor Mordatch, Ofir Nachum, Carolina Parada, Jodilyn Peralta, Emily Perez, Karl Pertsch, Jornell Quiambao, Kanishka Rao, Michael Ryoo, Grecia Salazar, Pannag Sanketi, Kevin Sayed, Jaspiar Singh, Sumedh Sontakke, Austin Stone, Clayton Tan, Huong Tran, Vincent Vanhoucke, Steve Vega, Quan Vuong, Fei Xia, Ted Xiao, Peng Xu, Sichun Xu, Tianhe Yu, and Brianna Zitkovich.
\newblock Rt-1: Robotics transformer for real-world control at scale, 2023.

\bibitem[Brown et~al.(2020)Brown, Mann, Ryder, Subbiah, Kaplan, Dhariwal, Neelakantan, Shyam, Sastry, Askell, Agarwal, Herbert-Voss, Krueger, Henighan, Child, Ramesh, Ziegler, Wu, Winter, Hesse, Chen, Sigler, Litwin, Gray, Chess, Clark, Berner, McCandlish, Radford, Sutskever, and Amodei]{gpt3.5}
Tom~B. Brown, Benjamin Mann, Nick Ryder, Melanie Subbiah, Jared Kaplan, Prafulla Dhariwal, Arvind Neelakantan, Pranav Shyam, Girish Sastry, Amanda Askell, Sandhini Agarwal, Ariel Herbert-Voss, Gretchen Krueger, Tom Henighan, Rewon Child, Aditya Ramesh, Daniel~M. Ziegler, Jeffrey Wu, Clemens Winter, Christopher Hesse, Mark Chen, Eric Sigler, Mateusz Litwin, Scott Gray, Benjamin Chess, Jack Clark, Christopher Berner, Sam McCandlish, Alec Radford, Ilya Sutskever, and Dario Amodei.
\newblock Language models are few-shot learners, 2020.

\bibitem[Chen et~al.(2023{\natexlab{a}})Chen, Dohan, and So]{chen2023evoprompting}
Angelica Chen, David~M. Dohan, and David~R. So.
\newblock Evoprompting: Language models for code-level neural architecture search, 2023{\natexlab{a}}.

\bibitem[Chen et~al.(2021{\natexlab{a}})Chen, Tworek, Jun, Yuan, de~Oliveira~Pinto, Kaplan, Edwards, Burda, Joseph, Brockman, Ray, Puri, Krueger, Petrov, Khlaaf, Sastry, Mishkin, Chan, Gray, Ryder, Pavlov, Power, Kaiser, Bavarian, Winter, Tillet, Such, Cummings, Plappert, Chantzis, Barnes, Herbert-Voss, Guss, Nichol, Paino, Tezak, Tang, Babuschkin, Balaji, Jain, Saunders, Hesse, Carr, Leike, Achiam, Misra, Morikawa, Radford, Knight, Brundage, Murati, Mayer, Welinder, McGrew, Amodei, McCandlish, Sutskever, and Zaremba]{chen2021evaluating}
Mark Chen, Jerry Tworek, Heewoo Jun, Qiming Yuan, Henrique~Ponde de~Oliveira~Pinto, Jared Kaplan, Harri Edwards, Yuri Burda, Nicholas Joseph, Greg Brockman, Alex Ray, Raul Puri, Gretchen Krueger, Michael Petrov, Heidy Khlaaf, Girish Sastry, Pamela Mishkin, Brooke Chan, Scott Gray, Nick Ryder, Mikhail Pavlov, Alethea Power, Lukasz Kaiser, Mohammad Bavarian, Clemens Winter, Philippe Tillet, Felipe~Petroski Such, Dave Cummings, Matthias Plappert, Fotios Chantzis, Elizabeth Barnes, Ariel Herbert-Voss, William~Hebgen Guss, Alex Nichol, Alex Paino, Nikolas Tezak, Jie Tang, Igor Babuschkin, Suchir Balaji, Shantanu Jain, William Saunders, Christopher Hesse, Andrew~N. Carr, Jan Leike, Josh Achiam, Vedant Misra, Evan Morikawa, Alec Radford, Matthew Knight, Miles Brundage, Mira Murati, Katie Mayer, Peter Welinder, Bob McGrew, Dario Amodei, Sam McCandlish, Ilya Sutskever, and Wojciech Zaremba.
\newblock Evaluating large language models trained on code, 2021{\natexlab{a}}.

\bibitem[Chen et~al.(2021{\natexlab{b}})Chen, Tworek, Jun, Yuan, Pinto, Kaplan, Edwards, Burda, Joseph, Brockman, et~al.]{codex}
Mark Chen, Jerry Tworek, Heewoo Jun, Qiming Yuan, Henrique Ponde de~Oliveira Pinto, Jared Kaplan, Harri Edwards, Yuri Burda, Nicholas Joseph, Greg Brockman, et~al.
\newblock {Evaluating Large Language Models Trained on Code}, 2021{\natexlab{b}}.
\newblock \textit{arXiv preprint arXiv:2107.03374.} \url{https://arxiv.org/abs/2107.03374}.

\bibitem[Chen et~al.(2023{\natexlab{b}})Chen, Lin, Sch\"{a}rli, and Zhou]{chen2023teaching}
Xinyun Chen, Maxwell Lin, Nathanael Sch\"{a}rli, and Denny Zhou.
\newblock {Teaching Large Language Models to Self-Debug}, 2023{\natexlab{b}}.
\newblock \textit{arXiv preprint arXiv:2304.05128.} \url{https://arxiv.org/abs/2304.05128}.

\bibitem[Chowdhery et~al.(2022)Chowdhery, Narang, Devlin, Bosma, Mishra, Roberts, Barham, Chung, Sutton, Gehrmann, et~al.]{chowdhery2022palm}
Aakanksha Chowdhery, Sharan Narang, Jacob Devlin, Maarten Bosma, Gaurav Mishra, Adam Roberts, Paul Barham, Hyung~Won Chung, Charles Sutton, Sebastian Gehrmann, et~al.
\newblock {PaLM: Scaling Language Modeling with Pathways}, 2022.
\newblock \textit{arXiv preprint arXiv:2204.02311.} \url{https://arxiv.org/abs/2204.02311}.

\bibitem[Ding et~al.(2023)Ding, Zhang, Paxton, and Zhang]{ding2023task}
Yan Ding, Xiaohan Zhang, Chris Paxton, and Shiqi Zhang.
\newblock Task and motion planning with large language models for object rearrangement, 2023.

\bibitem[Driess et~al.(2023)Driess, Xia, Sajjadi, Lynch, Chowdhery, Ichter, Wahid, Tompson, Vuong, Yu, Huang, Chebotar, Sermanet, Duckworth, Levine, Vanhoucke, Hausman, Toussaint, Greff, Zeng, Mordatch, and Florence]{palme}
Danny Driess, Fei Xia, Mehdi S.~M. Sajjadi, Corey Lynch, Aakanksha Chowdhery, Brian Ichter, Ayzaan Wahid, Jonathan Tompson, Quan Vuong, Tianhe Yu, Wenlong Huang, Yevgen Chebotar, Pierre Sermanet, Daniel Duckworth, Sergey Levine, Vincent Vanhoucke, Karol Hausman, Marc Toussaint, Klaus Greff, Andy Zeng, Igor Mordatch, and Pete Florence.
\newblock Palm-e: An embodied multimodal language model, 2023.

\bibitem[Fried et~al.(2023)Fried, Aghajanyan, Lin, Wang, Wallace, Shi, Zhong, Yih, Zettlemoyer, and Lewis]{fried2022incoder}
Daniel Fried, Armen Aghajanyan, Jessy Lin, Sida Wang, Eric Wallace, Freda Shi, Ruiqi Zhong, Wen-tau Yih, Luke Zettlemoyer, and Mike Lewis.
\newblock {InCoder: A generative model for code infilling and synthesis}.
\newblock In \emph{International Conference on Learning Representations}, 2023.

\bibitem[Guhur et~al.(2022)Guhur, Chen, Garcia, Tapaswi, Laptev, and Schmid]{guhur2022instruction}
Pierre-Louis Guhur, Shizhe Chen, Ricardo Garcia, Makarand Tapaswi, Ivan Laptev, and Cordelia Schmid.
\newblock Instruction-driven history-aware policies for robotic manipulations, 2022.

\bibitem[Guo et~al.(2024)Guo, Wang, Guo, Li, Song, Tan, Liu, Bian, and Yang]{guo2023connecting}
Qingyan Guo, Rui Wang, Junliang Guo, Bei Li, Kaitao Song, Xu~Tan, Guoqing Liu, Jiang Bian, and Yujiu Yang.
\newblock Connecting large language models with evolutionary algorithms yields powerful prompt optimizers, 2024.

\bibitem[Hendrycks et~al.(2021)Hendrycks, Basart, Kadavath, Mazeika, Arora, Guo, Burns, Puranik, He, Song, and Steinhardt]{hendrycksapps2021}
Dan Hendrycks, Steven Basart, Saurav Kadavath, Mantas Mazeika, Akul Arora, Ethan Guo, Collin Burns, Samir Puranik, Horace He, Dawn Song, and Jacob Steinhardt.
\newblock {Measuring Coding Challenge Competence With APPS}.
\newblock In \emph{Advances in Neural Information Processing Systems}, 2021.

\bibitem[Huang et~al.(2023)Huang, Jiang, Dong, Qiao, Gao, and Li]{huang2023instruct2act}
Siyuan Huang, Zhengkai Jiang, Hao Dong, Yu~Qiao, Peng Gao, and Hongsheng Li.
\newblock Instruct2act: Mapping multi-modality instructions to robotic actions with large language model, 2023.

\bibitem[Jang et~al.(2022)Jang, Irpan, Khansari, Kappler, Ebert, Lynch, Levine, and Finn]{jang2022bc}
Eric Jang, Alex Irpan, Mohi Khansari, Daniel Kappler, Frederik Ebert, Corey Lynch, Sergey Levine, and Chelsea Finn.
\newblock Bc-z: Zero-shot task generalization with robotic imitation learning, 2022.

\bibitem[Jiang et~al.(2023)Jiang, Gupta, Zhang, Wang, Dou, Chen, Fei-Fei, Anandkumar, Zhu, and Fan]{jiang2022vima}
Yunfan Jiang, Agrim Gupta, Zichen Zhang, Guanzhi Wang, Yongqiang Dou, Yanjun Chen, Li~Fei-Fei, Anima Anandkumar, Yuke Zhu, and Linxi Fan.
\newblock Vima: General robot manipulation with multimodal prompts, 2023.

\bibitem[Le et~al.(2022)Le, Wang, Gotmare, Savarese, and Hoi]{codeRL}
Hung Le, Yue Wang, Akhilesh~Deepak Gotmare, Silvio Savarese, and Steven Chu~Hong Hoi.
\newblock {CodeRL: Mastering Code Generation through Pretrained Models and Deep Reinforcement Learning}.
\newblock In \emph{Advances in Neural Information Processing Systems}, 2022.

\bibitem[Lehman et~al.(2022)Lehman, Gordon, Jain, Ndousse, Yeh, and Stanley]{lehman2022evolution}
Joel Lehman, Jonathan Gordon, Shawn Jain, Kamal Ndousse, Cathy Yeh, and Kenneth~O. Stanley.
\newblock Evolution through large models, 2022.

\bibitem[Li et~al.(2023{\natexlab{a}})Li, Allal, Zi, Muennighoff, Kocetkov, Mou, Marone, Akiki, Li, Chim, et~al.]{li2023starcoder}
Raymond Li, Loubna~Ben Allal, Yangtian Zi, Niklas Muennighoff, Denis Kocetkov, Chenghao Mou, Marc Marone, Christopher Akiki, Jia Li, Jenny Chim, et~al.
\newblock {StarCoder: may the source be with you!}, 2023{\natexlab{a}}.
\newblock \textit{arXiv preprint arXiv:2305.06161.} \url{https://arxiv.org/abs/2305.06161}.

\bibitem[Li et~al.(2023{\natexlab{b}})Li, Allal, Zi, Muennighoff, Kocetkov, Mou, Marone, Akiki, Li, Chim, et~al.]{starcoder}
Raymond Li, Loubna~Ben Allal, Yangtian Zi, Niklas Muennighoff, Denis Kocetkov, Chenghao Mou, Marc Marone, Christopher Akiki, Jia Li, Jenny Chim, et~al.
\newblock Starcoder: may the source be with you!
\newblock \emph{arXiv preprint arXiv:2305.06161}, 2023{\natexlab{b}}.

\bibitem[Li et~al.(2022)Li, Choi, Chung, Kushman, Schrittwieser, Leblond, Eccles, Keeling, Gimeno, Lago, et~al.]{alphacode}
Yujia Li, David Choi, Junyoung Chung, Nate Kushman, Julian Schrittwieser, Rémi Leblond, Tom Eccles, James Keeling, Felix Gimeno, Agustin~Dal Lago, et~al.
\newblock {Competition-level code generation with AlphaCode}.
\newblock \emph{Science}, 378\penalty0 (6624):\penalty0 1092--1097, 2022.
\newblock \doi{10.1126/science.abq1158}.

\bibitem[Liang et~al.(2023)Liang, Huang, Xia, Xu, Hausman, Ichter, Florence, and Zeng]{liang2023code}
Jacky Liang, Wenlong Huang, Fei Xia, Peng Xu, Karol Hausman, Brian Ichter, Pete Florence, and Andy Zeng.
\newblock Code as policies: Language model programs for embodied control, 2023.

\bibitem[Lin et~al.(2023)Lin, Agia, Migimatsu, Pavone, and Bohg]{lin2023text2motion}
Kevin Lin, Christopher Agia, Toki Migimatsu, Marco Pavone, and Jeannette Bohg.
\newblock Text2motion: from natural language instructions to feasible plans.
\newblock \emph{Autonomous Robots}, 47\penalty0 (8):\penalty0 1345–1365, November 2023.
\newblock ISSN 1573-7527.
\newblock \doi{10.1007/s10514-023-10131-7}.
\newblock URL \url{http://dx.doi.org/10.1007/s10514-023-10131-7}.

\bibitem[Liu et~al.(2023)Liu, Jiang, Zhang, Liu, Zhang, Biswas, and Stone]{liu2023llm+}
Bo~Liu, Yuqian Jiang, Xiaohan Zhang, Qiang Liu, Shiqi Zhang, Joydeep Biswas, and Peter Stone.
\newblock Llm+p: Empowering large language models with optimal planning proficiency, 2023.

\bibitem[Luo et~al.(2023)Luo, Xu, Zhao, Sun, Geng, Hu, Tao, Ma, Lin, and Jiang]{wizardcoder}
Ziyang Luo, Can Xu, Pu~Zhao, Qingfeng Sun, Xiubo Geng, Wenxiang Hu, Chongyang Tao, Jing Ma, Qingwei Lin, and Daxin Jiang.
\newblock Wizardcoder: Empowering code large language models with evol-instruct, 2023.

\bibitem[Ma et~al.(2023)Ma, Liang, Wang, Huang, Bastani, Jayaraman, Zhu, Fan, and Anandkumar]{eureka}
Yecheng~Jason Ma, William Liang, Guanzhi Wang, De-An Huang, Osbert Bastani, Dinesh Jayaraman, Yuke Zhu, Linxi Fan, and Anima Anandkumar.
\newblock Eureka: Human-level reward design via coding large language models, 2023.

\bibitem[Madaan et~al.(2023)Madaan, Tandon, Gupta, Hallinan, Gao, Wiegreffe, Alon, Dziri, Prabhumoye, Yang, et~al.]{madaan2023selfrefine}
Aman Madaan, Niket Tandon, Prakhar Gupta, Skyler Hallinan, Luyu Gao, Sarah Wiegreffe, Uri Alon, Nouha Dziri, Shrimai Prabhumoye, Yiming Yang, et~al.
\newblock {Self-Refine: Iterative Refinement with Self-Feedback}, 2023.
\newblock \textit{arXiv preprint arXiv:2303.17651.} \url{https://arxiv.org/abs/2303.17651}.

\bibitem[Makoviychuk et~al.(2021)Makoviychuk, Wawrzyniak, Guo, Lu, Storey, Macklin, Hoeller, Rudin, Allshire, Handa, and State]{isaac}
Viktor Makoviychuk, Lukasz Wawrzyniak, Yunrong Guo, Michelle Lu, Kier Storey, Miles Macklin, David Hoeller, Nikita Rudin, Arthur Allshire, Ankur Handa, and Gavriel State.
\newblock Isaac gym: High performance gpu-based physics simulation for robot learning, 2021.

\bibitem[Mu et~al.(2024)Mu, Chen, Zhang, Chen, Yu, Ge, Chen, Liang, Hu, Tao, Sun, Yu, Yang, Shao, Wang, Dai, Qiao, Ding, and Luo]{robocodex}
Yao Mu, Junting Chen, Qinglong Zhang, Shoufa Chen, Qiaojun Yu, Chongjian Ge, Runjian Chen, Zhixuan Liang, Mengkang Hu, Chaofan Tao, Peize Sun, Haibao Yu, Chao Yang, Wenqi Shao, Wenhai Wang, Jifeng Dai, Yu~Qiao, Mingyu Ding, and Ping Luo.
\newblock Robocodex: Multimodal code generation for robotic behavior synthesis, 2024.

\bibitem[Nasir et~al.(2023)Nasir, Earle, Togelius, James, and Cleghorn]{nasir2023llmatic}
Muhammad~U. Nasir, Sam Earle, Julian Togelius, Steven James, and Christopher Cleghorn.
\newblock Llmatic: Neural architecture search via large language models and quality diversity optimization, 2023.

\bibitem[Nijkamp et~al.(2023)Nijkamp, Pang, Hayashi, Tu, Wang, Zhou, Savarese, and Xiong]{codegen}
Erik Nijkamp, Bo~Pang, Hiroaki Hayashi, Lifu Tu, Huan Wang, Yingbo Zhou, Silvio Savarese, and Caiming Xiong.
\newblock {CodeGen: An Open Large Language Model for Code with Multi-Turn Program Synthesis}.
\newblock In \emph{International Conference on Learning Representations}, 2023.

\bibitem[OpenAI(2023)]{openai2023gpt4}
OpenAI.
\newblock {GPT-4 Technical Report}, 2023.
\newblock \textit{arXiv preprint arXiv:2303.08774.} \url{https://arxiv.org/abs/2303.08774}.

\bibitem[Roziere et~al.(2023)Roziere, Gehring, Gloeckle, Sootla, Gat, Tan, Adi, Liu, Remez, Rapin, et~al.]{codellama}
Baptiste Roziere, Jonas Gehring, Fabian Gloeckle, Sten Sootla, Itai Gat, Xiaoqing~Ellen Tan, Yossi Adi, Jingyu Liu, Tal Remez, J{\'e}r{\'e}my Rapin, et~al.
\newblock Code llama: Open foundation models for code.
\newblock \emph{arXiv preprint arXiv:2308.12950}, 2023.

\bibitem[Shinn et~al.(2023)Shinn, Cassano, Berman, Gopinath, Narasimhan, and Yao]{shinn2023reflexion}
Noah Shinn, Federico Cassano, Edward Berman, Ashwin Gopinath, Karthik Narasimhan, and Shunyu Yao.
\newblock Reflexion: Language agents with verbal reinforcement learning, 2023.

\bibitem[Shridhar et~al.(2022)Shridhar, Manuelli, and Fox]{shridhar2022perceiver}
Mohit Shridhar, Lucas Manuelli, and Dieter Fox.
\newblock Perceiver-actor: A multi-task transformer for robotic manipulation, 2022.

\bibitem[Silva et~al.(2022)Silva, Moorman, Silva, Zaidi, Gopalan, and Gombolay]{silva2021lancon}
Andrew Silva, Nina Moorman, William Silva, Zulfiqar Zaidi, Nakul Gopalan, and Matthew Gombolay.
\newblock Lancon-learn: Learning with language to enable generalization in multi-task manipulation.
\newblock \emph{IEEE Robotics and Automation Letters}, 7\penalty0 (2):\penalty0 1635--1642, 2022.
\newblock \doi{10.1109/LRA.2021.3139667}.

\bibitem[Silver et~al.(2023)Silver, Dan, Srinivas, Tenenbaum, Kaelbling, and Katz]{silver2023generalized}
Tom Silver, Soham Dan, Kavitha Srinivas, Joshua~B. Tenenbaum, Leslie~Pack Kaelbling, and Michael Katz.
\newblock Generalized planning in pddl domains with pretrained large language models, 2023.

\bibitem[Singh et~al.(2022)Singh, Blukis, Mousavian, Goyal, Xu, Tremblay, Fox, Thomason, and Garg]{singh2023progprompt}
Ishika Singh, Valts Blukis, Arsalan Mousavian, Ankit Goyal, Danfei Xu, Jonathan Tremblay, Dieter Fox, Jesse Thomason, and Animesh Garg.
\newblock Progprompt: Generating situated robot task plans using large language models, 2022.

\bibitem[Touvron et~al.(2023)Touvron, Lavril, Izacard, Martinet, Lachaux, Lacroix, Rozi{\`e}re, Goyal, Hambro, Azhar, et~al.]{touvron2023llama}
Hugo Touvron, Thibaut Lavril, Gautier Izacard, Xavier Martinet, Marie-Anne Lachaux, Timoth{\'e}e Lacroix, Baptiste Rozi{\`e}re, Naman Goyal, Eric Hambro, Faisal Azhar, et~al.
\newblock {Llama: Open and efficient foundation language models}, 2023.
\newblock \textit{arXiv preprint arXiv:2302.13971.} \url{https://arxiv.org/abs/2302.13971}.

\bibitem[Wang et~al.(2023{\natexlab{a}})Wang, Xie, Jiang, Mandlekar, Xiao, Zhu, Fan, and Anandkumar]{wang2023voyager}
Guanzhi Wang, Yuqi Xie, Yunfan Jiang, Ajay Mandlekar, Chaowei Xiao, Yuke Zhu, Linxi Fan, and Anima Anandkumar.
\newblock Voyager: An open-ended embodied agent with large language models.
\newblock In \emph{NeurIPS 2023 Foundation Models for Decision Making Workshop}, 2023{\natexlab{a}}.
\newblock URL \url{https://openreview.net/forum?id=P8E4Br72j3}.

\bibitem[Wang et~al.(2023{\natexlab{b}})Wang, Gonzalez-Pumariega, Sharma, and Choudhury]{wang2023demo2code}
Huaxiaoyue Wang, Gonzalo Gonzalez-Pumariega, Yash Sharma, and Sanjiban Choudhury.
\newblock Demo2code: From summarizing demonstrations to synthesizing code via extended chain-of-thought, 2023{\natexlab{b}}.

\bibitem[Wang et~al.(2020)Wang, Wei, Dong, Bao, Yang, and Zhou]{minilm}
Wenhui Wang, Furu Wei, Li~Dong, Hangbo Bao, Nan Yang, and Ming Zhou.
\newblock Minilm: Deep self-attention distillation for task-agnostic compression of pre-trained transformers, 2020.

\bibitem[Wang et~al.(2021)Wang, Bao, Huang, Dong, and Wei]{minilmv2}
Wenhui Wang, Hangbo Bao, Shaohan Huang, Li~Dong, and Furu Wei.
\newblock Minilmv2: Multi-head self-attention relation distillation for compressing pretrained transformers, 2021.

\bibitem[Xie et~al.(2023)Xie, Yu, Zhu, Bai, Gong, and Soh]{xie2023translating}
Yaqi Xie, Chen Yu, Tongyao Zhu, Jinbin Bai, Ze~Gong, and Harold Soh.
\newblock Translating natural language to planning goals with large-language models, 2023.

\bibitem[Yu et~al.(2023)Yu, Gileadi, Fu, Kirmani, Lee, Arenas, Chiang, Erez, Hasenclever, Humplik, Ichter, Xiao, Xu, Zeng, Zhang, Heess, Sadigh, Tan, Tassa, and Xia]{yu2023language}
Wenhao Yu, Nimrod Gileadi, Chuyuan Fu, Sean Kirmani, Kuang-Huei Lee, Montse~Gonzalez Arenas, Hao-Tien~Lewis Chiang, Tom Erez, Leonard Hasenclever, Jan Humplik, Brian Ichter, Ted Xiao, Peng Xu, Andy Zeng, Tingnan Zhang, Nicolas Heess, Dorsa Sadigh, Jie Tan, Yuval Tassa, and Fei Xia.
\newblock Language to rewards for robotic skill synthesis, 2023.

\bibitem[Zhang et~al.(2023)Zhang, Li, Li, Li, and Jin]{zhang2023selfedit}
Kechi Zhang, Zhuo Li, Jia Li, Ge~Li, and Zhi Jin.
\newblock {Self-Edit: Fault-Aware Code Editor for Code Generation}, 2023.
\newblock \textit{arXiv preprint arXiv:2305.04087.} \url{https://arxiv.org/abs/2305.04087}.

\bibitem[Zhang \& Chai(2021)Zhang and Chai]{zhang2021hierarchical}
Yichi Zhang and Joyce Chai.
\newblock Hierarchical task learning from language instructions with unified transformers and self-monitoring.
\newblock \emph{arXiv preprint arXiv:2106.03427}, 2021.

\end{thebibliography}
\bibliographystyle{iclr2024_conference}

\newpage
\appendix
\iffalse
In the appendix, we supplement the content that was not elaborated in the main text due to pages limitations.
\cref{sec:impact} shows the broader impacts of the paper.
\cref{sec:dofs} shows the algorithms of our framework and Searcher.
\cref{sec:dofs} introduces the concept of Degrees of Freedom (DOF) and detail the DOF properties for entities. 
\cref{sec:test_space} is dedicated to presenting a test space for evaluating the effectiveness of our baseline framework across the varied entities. 
\cref{sec:action_space} highlights the initial action space of our framework. 
\cref{sec:code_evaluator} compares the pass rates as determined by the Image Evaluator and the Code Evaluator.
\fi

%%%%%%%%%%%%%%%%%%%%%%%%%%%%%%%%%%%%%%%%%%%%%%%%%%%%%%%%%%%%%%%%
\section{Broader impacts}
\label{sec:impact}
%%%%%%%%%%%%%%%%%%%%%%%%%%%%%%%%%%%%%%%%%%%%%%%%%%%%%%%%%%%%%%%%

\paragraph{Positive societal impacts.} We propose a method for robotic manipulation, which will be helpful for users to manipulate robotics.

\paragraph{negative societal impacts.} This approach may be abused of other potential robotic manipulation applications.

%%%%%%%%%%%%%%%%%%%%%%%%%%%%%%%%%%%%%%%%%%%%%%%%%%%%%%%%%%%%%%%%
\section{Algorithm}
\label{sec:alg}
%%%%%%%%%%%%%%%%%%%%%%%%%%%%%%%%%%%%%%%%%%%%%%%%%%%%%%%%%%%%%%%%
The algorithms of our framework and Searcher are shown below.

\begin{figure}[h]
    \centering
    \begin{minipage}{.48\textwidth}
        \input{algorithms/framework}
        \label{alg:framework}
    \end{minipage}%
    \hfill
    \begin{minipage}{.48\textwidth}
        \input{algorithms/searcher}
        \label{alg:searcher}
    \end{minipage}%
\end{figure}

%%%%%%%%%%%%%%%%%%%%%%%%%%%%%%%%%%%%%%%%%%%%%%%%%%%%%%%%%%%%%%%%
\section{DOFs}
\label{sec:dofs}
%%%%%%%%%%%%%%%%%%%%%%%%%%%%%%%%%%%%%%%%%%%%%%%%%%%%%%%%%%%%%%%%
In this section, we present the Degrees of Freedom (DOF) properties (\cref{sec:dof_properties}) and illustrate their movements (\cref{sec:dof_illustrations}).

%%%%%%%%%%%%%%%%%%%%%%%%%%%%%%%%%%%%%%%%%%%%%%%%%%%%%%%%%%%%%%%%
\subsection{DOFs Properties}
\label{sec:dof_properties}
%%%%%%%%%%%%%%%%%%%%%%%%%%%%%%%%%%%%%%%%%%%%%%%%%%%%%%%%%%%%%%%%

In this section, we present a collection of tables, each representing a unique environment: Human~(\cref{tab:dofs_human}), Ant~(\cref{tab:dofs_ant}), Cartpole~(\cref{tab:dofs_cartpole}), Sektion Cabinet~(\cref{tab:dofs_cabinet}), Franka Panda~(\cref{tab:dofs_panda}), Kinova Cabinet ~(\cref{tab:dofs_cabinet}), and Anymal~(\cref{tab:dofs_anymal}). These tables serve as a comprehensive guide, meticulously documenting the properties for each system, including Degrees of Freedom (DOF), Name, Type, Stiffness, Damping, Armature, and the constraints on motion as defined by Lower and Upper Limits. Through these tables, readers are afforded a clear understanding of the detailed properties of each environment.

\begin{table}[h]
\centering
\caption{Human DOF Properties}
\setlength{\tabcolsep}{0.5mm}
\begin{tabular}{cllrrrrrr}
\toprule
DOF & Name & Type & Stiffness & Damping & Armature & Limited? & Lower Limit & Upper Limit \\ 
\midrule
0 & abdomen\_z & Rotation & 20.00 & 5.00 & 0.02 & Yes & -0.79 & 0.79 \\
1 & abdomen\_y & Rotation & 20.00 & 5.00 & 0.01 & Yes & -1.31 & 0.52 \\
2 & abdomen\_x & Rotation & 10.00 & 5.00 & 0.01 & Yes & -0.61 & 0.61 \\
3 & right\_hip\_x & Rotation & 10.00 & 5.00 & 0.01 & Yes & -0.79 & 0.26 \\
4 & right\_hip\_z & Rotation & 10.00 & 5.00 & 0.01 & Yes & -1.05 & 0.61 \\
5 & right\_hip\_y & Rotation & 20.00 & 5.00 & 0.01 & Yes & -2.09 & 0.79 \\
6 & right\_knee & Rotation & 5.00 & 0.10 & 0.01 & Yes & -2.79 & 0.03 \\
7 & right\_ankle\_y & Rotation & 2.00 & 1.00 & 0.01 & Yes & -0.87 & 0.87 \\
8 & right\_ankle\_x & Rotation & 2.00 & 1.00 & 0.01 & Yes & -0.87 & 0.87 \\
9 & left\_hip\_x & Rotation & 10.00 & 5.00 & 0.01 & Yes & -0.79 & 0.26 \\
10 & left\_hip\_z & Rotation & 10.00 & 5.00 & 0.01 & Yes & -1.05 & 0.61 \\
11 & left\_hip\_y & Rotation & 20.00 & 5.00 & 0.01 & Yes & -2.09 & 0.79 \\
12 & left\_knee & Rotation & 5.00 & 0.10 & 0.01 & Yes & -2.79 & 0.03 \\
13 & left\_ankle\_y & Rotation & 2.00 & 1.00 & 0.01 & Yes & -0.87 & 0.87 \\
14 & left\_ankle\_x & Rotation & 2.00 & 1.00 & 0.01 & Yes & -0.87 & 0.87 \\
15 & right\_shoulder1 & Rotation & 10.00 & 5.00 & 0.01 & Yes & -1.57 & 1.22 \\
16 & right\_shoulder2 & Rotation & 10.00 & 5.00 & 0.01 & Yes & -1.57 & 1.22 \\
17 & right\_elbow & Rotation & 2.00 & 1.00 & 0.01 & Yes & -1.57 & 0.87 \\
18 & left\_shoulder1 & Rotation & 10.00 & 5.00 & 0.01 & Yes & -1.57 & 1.22 \\
19 & left\_shoulder2 & Rotation & 10.00 & 5.00 & 0.01 & Yes & -1.57 & 1.22 \\
20 & left\_elbow & Rotation & 2.00 & 1.00 & 0.01 & Yes & -1.57 & 0.87 \\ 
\bottomrule
\end{tabular}
\label{tab:dofs_human}
\end{table}

\begin{table}[h]
\centering
\caption{Ant DOF Properties}
\setlength{\tabcolsep}{1.2mm}
\begin{tabular}{cllrrrrrr}
\toprule
DOF & Name & Type & Stiffness & Damping & Armature & Limited? & Lower Limit & Upper Limit \\ 
\midrule
0 & hip\_1 & Rotation & 0.00 & 0.10 & 0.01 & Yes & -0.70 & 0.70 \\
1 & ankle\_1 & Rotation & 0.00 & 0.10 & 0.01 & Yes & 0.52 & 1.75 \\
2 & hip\_2 & Rotation & 0.00 & 0.10 & 0.01 & Yes & -0.70 & 0.70 \\
3 & ankle\_2 & Rotation & 0.00 & 0.10 & 0.01 & Yes & -1.75 & -0.52 \\
4 & hip\_3 & Rotation & 0.00 & 0.10 & 0.01 & Yes & -0.70 & 0.70 \\
5 & ankle\_3 & Rotation & 0.00 & 0.10 & 0.01 & Yes & -1.75 & -0.52 \\
6 & hip\_4 & Rotation & 0.00 & 0.10 & 0.01 & Yes & -0.70 & 0.70 \\
7 & ankle\_4 & Rotation & 0.00 & 0.10 & 0.01 & Yes & 0.52 & 1.75 \\ \bottomrule
\end{tabular}
\label{tab:dofs_ant}
\end{table}

\begin{table}[h]
\centering
\caption{Cartpole DOF Properties}
\setlength{\tabcolsep}{0.3mm}
\begin{tabular}{cllrrrrrr}
\toprule
DOF & Name               & Type         & Stiffness    & Damping & Armature & Limited? & Lower Limit & Upper Limit \\
\midrule
0   & slider\_to\_cart   & Translation  & 0.0          & 0.0     & 0.00     & Yes      & -4.00       & 4.00        \\
1   & cart\_to\_pole     & Rotation     & $3.4 \times 10^{38}$ & 0.0     & 0.00     & No       & -3.14       & 3.14        \\
\bottomrule
\end{tabular}
\label{tab:dofs_cartpole}
\end{table}

\begin{table}[h]
\centering
\caption{Sektion Cabinet DOF Properties}
\setlength{\tabcolsep}{0.3mm}
\begin{tabular}{cllrrrrrr}
\toprule
DOF & Name & Type & Stiffness & Damping & Armature & Limited? & Lower Limit & Upper Limit \\
\midrule
0 & door\_left & Rotation & 0.0 & 0.1 & 0.00 & Yes & -1.57 & 0.00 \\
1 & door\_right & Rotation & 0.0 & 0.1 & 0.00 & Yes & 0.00 & 1.57 \\
2 & drawer\_bottom & Translation & 0.0 & 0.0 & 0.00 & Yes & 0.00 & 0.40 \\
3 & drawer\_top & Translation & 0.0 & 0.0 & 0.00 & Yes & 0.00 & 0.40 \\
\bottomrule
\end{tabular}
\label{tab:dofs_cabinet}
\end{table}

\begin{table}[h]
\centering
\caption{Franka Panda DOF Properties}
\setlength{\tabcolsep}{1mm}
\begin{tabular}{cllrrrrrr}
\toprule
DOF & Name & Type & Stiffness & Damping & Armature & Limited? & Lower Limit & Upper Limit \\
\midrule
0 & joint\_1 & Rotation & 0.0 & 10.0 & 0.00 & Yes & -2.90 & 2.90 \\
1 & joint\_2 & Rotation & 0.0 & 10.0 & 0.00 & Yes & -1.76 & 1.76 \\
2 & joint\_3 & Rotation & 0.0 & 10.0 & 0.00 & Yes & -2.90 & 2.90 \\
3 & joint\_4 & Rotation & 0.0 & 10.0 & 0.00 & Yes & -3.07 & -0.07 \\
4 & joint\_5 & Rotation & 0.0 & 10.0 & 0.00 & Yes & -2.90 & 2.90 \\
5 & joint\_6 & Rotation & 0.0 & 10.0 & 0.00 & Yes & -0.02 & 3.14 \\
6 & joint\_7 & Rotation & 0.0 & 10.0 & 0.00 & Yes & -2.90 & 2.90 \\
7 & finger\_1 & Translation & 0.0 & 10.0 & 0.00 & Yes & 0.00 & 0.04 \\
8 & finger\_2 & Translation & 0.0 & 10.0 & 0.00 & Yes & 0.00 & 0.04 \\
\bottomrule
\end{tabular}
\label{tab:dofs_panda}
\end{table}

\begin{table}[h]
\centering
\caption{Kinova DOF Properties}
\setlength{\tabcolsep}{1mm}
\begin{tabular}{cllrrrrrr}
\toprule
DOF & Name & Type & Stiffness & Damping & Armature & Limited? & Lower Limit & Upper Limit \\
\midrule
0 & joint\_1 & Rotation & 0.0 & 0.0 & 0.00 & Yes & -3.14 & 3.14 \\
1 & joint\_2 & Rotation & 0.0 & 0.0 & 0.00 & Yes & -3.14 & 3.14 \\
2 & joint\_3 & Rotation & 0.0 & 0.0 & 0.00 & Yes & -3.14 & 3.14 \\
3 & joint\_4 & Rotation & 0.0 & 0.0 & 0.00 & Yes & -3.14 & 3.14 \\
4 & joint\_5 & Rotation & 0.0 & 0.0 & 0.00 & Yes & -3.14 & 3.14 \\
5 & joint\_6 & Rotation & 0.0 & 0.0 & 0.00 & Yes & -3.14 & 3.14 \\
\bottomrule
\end{tabular}
\label{tab:dofs_kinova}
\end{table}

\begin{table}[h]
\centering
\caption{Anymal DOF Properties}
\setlength{\tabcolsep}{1mm}
\begin{tabular}{cllrrrrrr}
\toprule
DOF & Name & Type & Stiffness & Damping & Armature & Limited? & Lower Limit & Upper Limit \\
\midrule
0 & LF\_HAA & Rotation & 0.0 & 0.0 & 0.00 & Yes & -3.14 & 3.14 \\
1 & LF\_HFE & Rotation & 0.0 & 0.0 & 0.00 & Yes & -3.14 & 3.14 \\
2 & LF\_KFE & Rotation & 0.0 & 0.0 & 0.00 & Yes & -3.14 & 3.14 \\
3 & LH\_HAA & Rotation & 0.0 & 0.0 & 0.00 & Yes & -3.14 & 3.14 \\
4 & LH\_HFE & Rotation & 0.0 & 0.0 & 0.00 & Yes & -3.14 & 3.14 \\
5 & LH\_KFE & Rotation & 0.0 & 0.0 & 0.00 & Yes & -3.14 & 3.14 \\
6 & RF\_HAA & Rotation & 0.0 & 0.0 & 0.00 & Yes & -3.14 & 3.14 \\
7 & RF\_HFE & Rotation & 0.0 & 0.0 & 0.00 & Yes & -3.14 & 3.14 \\
8 & RF\_KFE & Rotation & 0.0 & 0.0 & 0.00 & Yes & -3.14 & 3.14 \\
9 & RH\_HAA & Rotation & 0.0 & 0.0 & 0.00 & Yes & -3.14 & 3.14 \\
10 & RH\_HFE & Rotation & 0.0 & 0.0 & 0.00 & Yes & -3.14 & 3.14 \\
11 & RH\_KFE & Rotation & 0.0 & 0.0 & 0.00 & Yes & -3.14 & 3.14 \\
\bottomrule
\end{tabular}
\label{tab:dofs_anymal}
\end{table}

%%%%%%%%%%%%%%%%%%%%%%%%%%%%%%%%%%%%%%%%%%%%%%%%%%%%%%%%%%%%%%%
\subsection{Property Distribution}
\label{sec:property_distribution}
%%%%%%%%%%%%%%%%%%%%%%%%%%%%%%%%%%%%%%%%%%%%%%%%%%%%%%%%%%%%%%%%

\begin{figure*}[ht]
\centering 
\includegraphics[width=\linewidth, trim=0 0 0 0, clip]{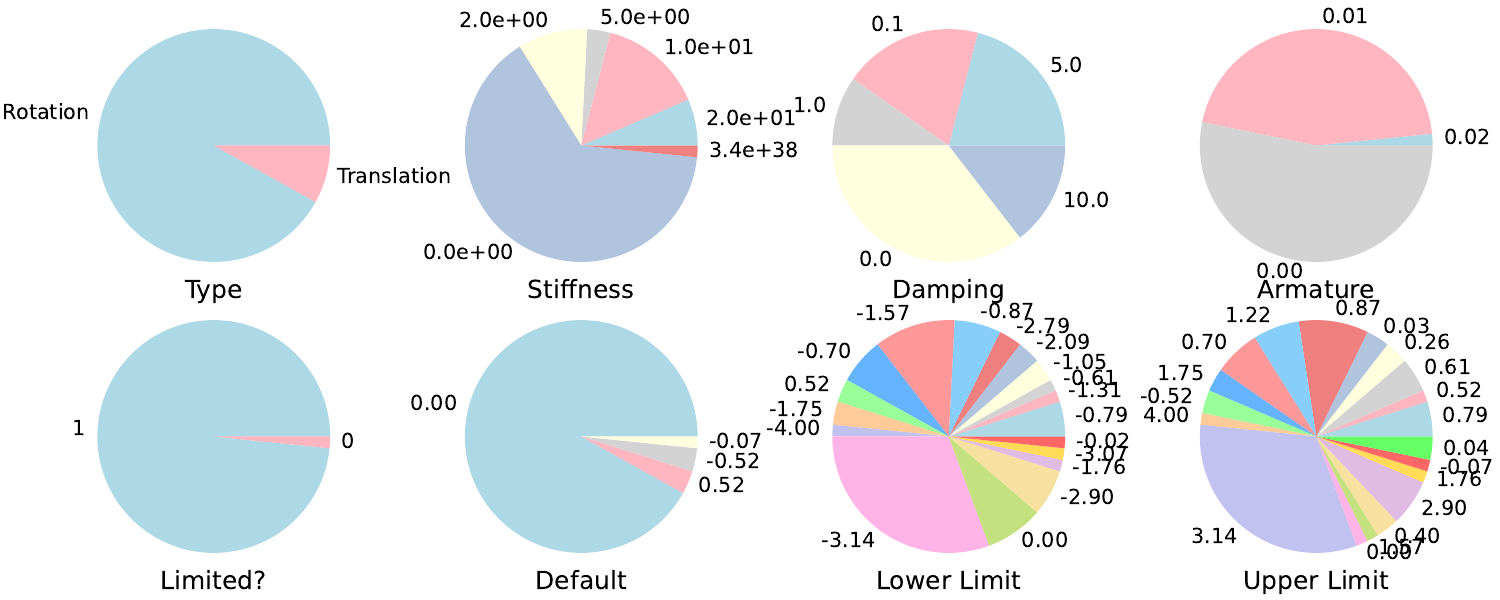}
\caption{The distribution of different properties across the different entities. The categories include \textit{Type, Stiffness, Damping, Armature, Limited?, Default, Lower Limit, and Upper Limit}. Each chart represents the relative frequency of parameter values, providing insights into the common settings used in the simulations.}
\label{fig:properties}
\end{figure*}

In \cref{fig:properties}, we present the parameter distributions for the entities. The \textit{Type} chart shows the type of movement of the DOFs. \textit{Stiffness} and \textit{Damping} parameters exhibit a wide range, with a notable preference for zero \textit{Stiffness}. The \textit{Armature} values are largely set to zero, suggesting minimal inertial effects are modeled in most cases. A stark majority of the parameters are \textit{Limited}, highlighting boundary conditions in the simulations. \textit{Default} positions are typically centered at zero, which is standard for simulations starting at a neutral state. The \textit{Lower Limit} and \textit{Upper Limit} distributions emphasize the angular constraints imposed on rotational movements, with significant clustering around $\pm 3.14$, suggestive of constraints set to the range of a full circle. These visualizations collectively allow for a quick assessment of the models' configuration tendencies, which can inform future model adjustments and comparative analysis.

%%%%%%%%%%%%%%%%%%%%%%%%%%%%%%%%%%%%%%%%%%%%%%%%%%%%%%%%%%%%%%%%
\subsection{DOFs Illustrations}
\label{sec:dof_illustrations}
\begin{figure}[ht]
    \centering
    \begin{minipage}{.33\linewidth}
        \includegraphics[width=\linewidth, trim=0 0 0 0, clip]{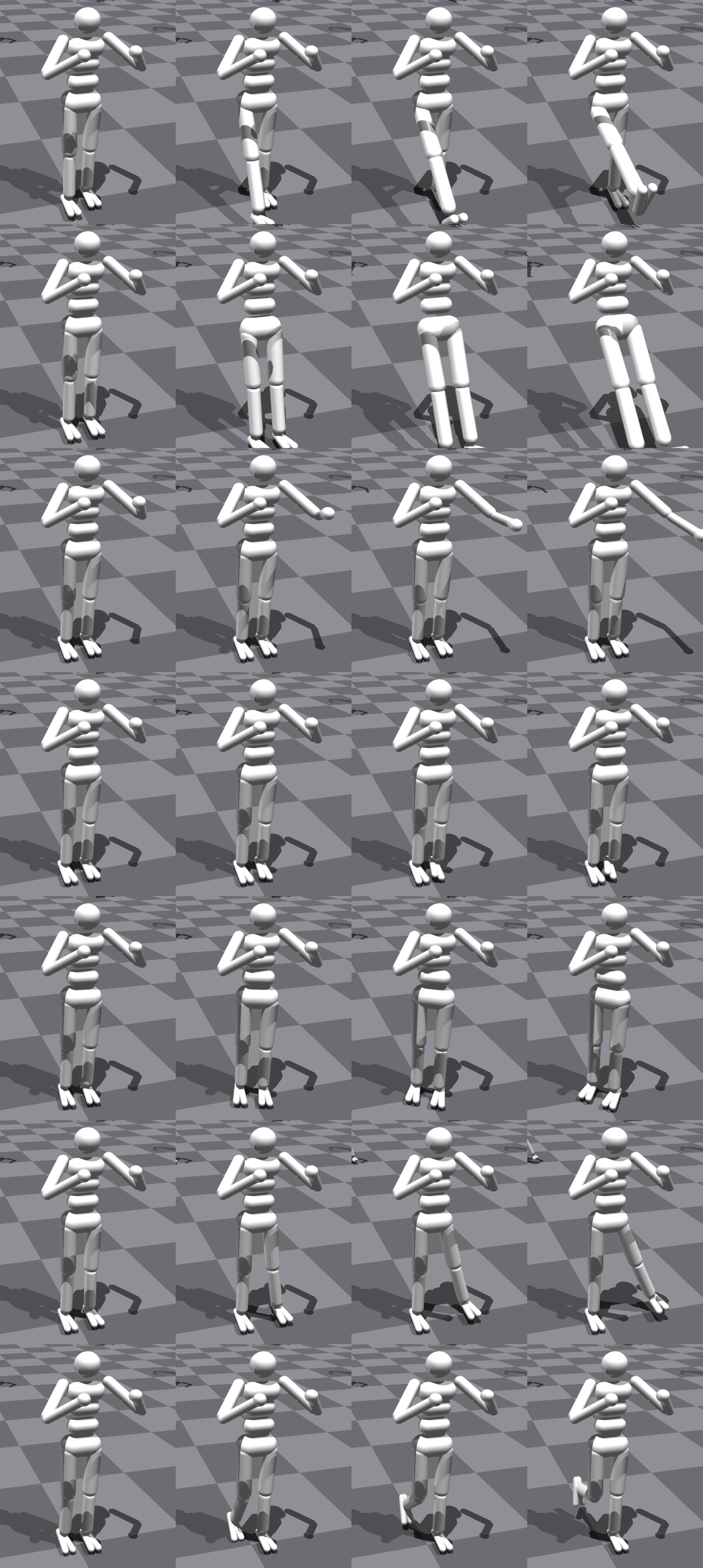}
        \subcaption{DOFs 0-6.}
    \end{minipage}%
    \begin{minipage}{.33\linewidth}
        \includegraphics[width=\linewidth, trim=0 0 0 0, clip]{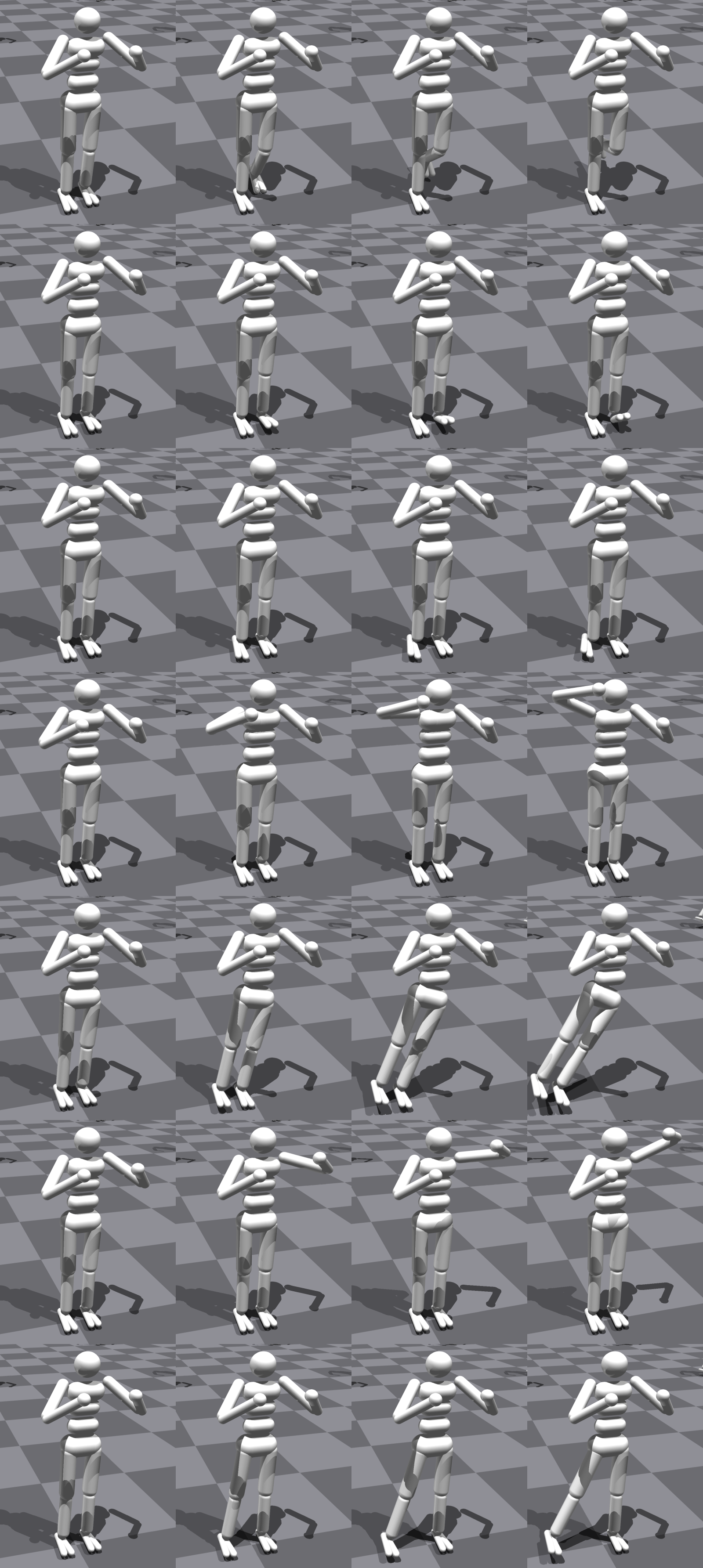}
        \subcaption{DOFs 7-13.}
    \end{minipage}%
    \begin{minipage}{.33\linewidth}
        \includegraphics[width=\linewidth, trim=0 0 0 0, clip]{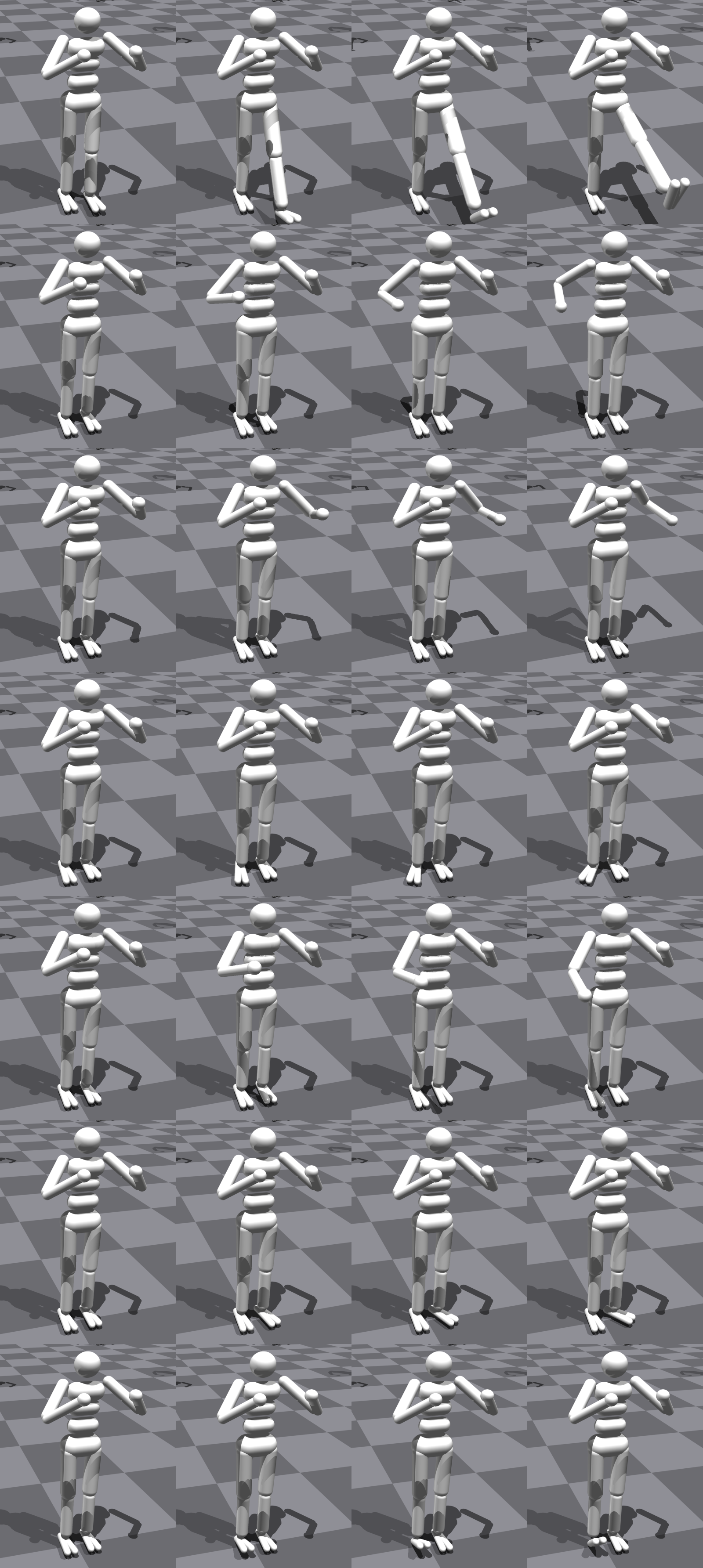}
        \subcaption{DOFs 14-21.}
    \end{minipage}%
    \caption{DOFs of Human.}
\label{fig:dofs_human}
\end{figure}

\begin{figure}[ht]
    \centering
    \begin{minipage}{.49\linewidth}
        \includegraphics[width=\linewidth, trim=0 0 0 0, clip]{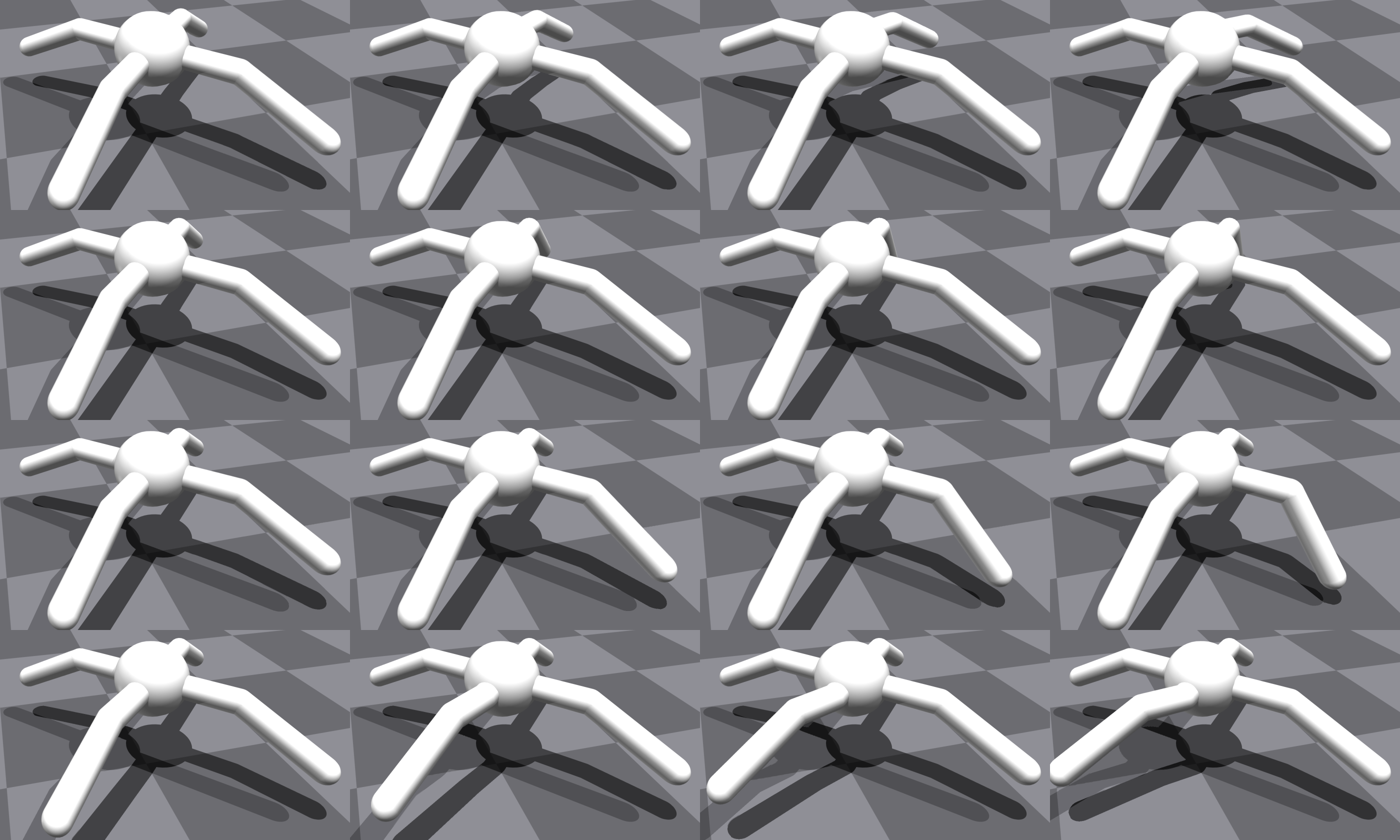}
        \subcaption{DOFs 0-3.}
    \end{minipage}%
    \begin{minipage}{.49\linewidth}
        \includegraphics[width=\linewidth, trim=0 0 0 0, clip]{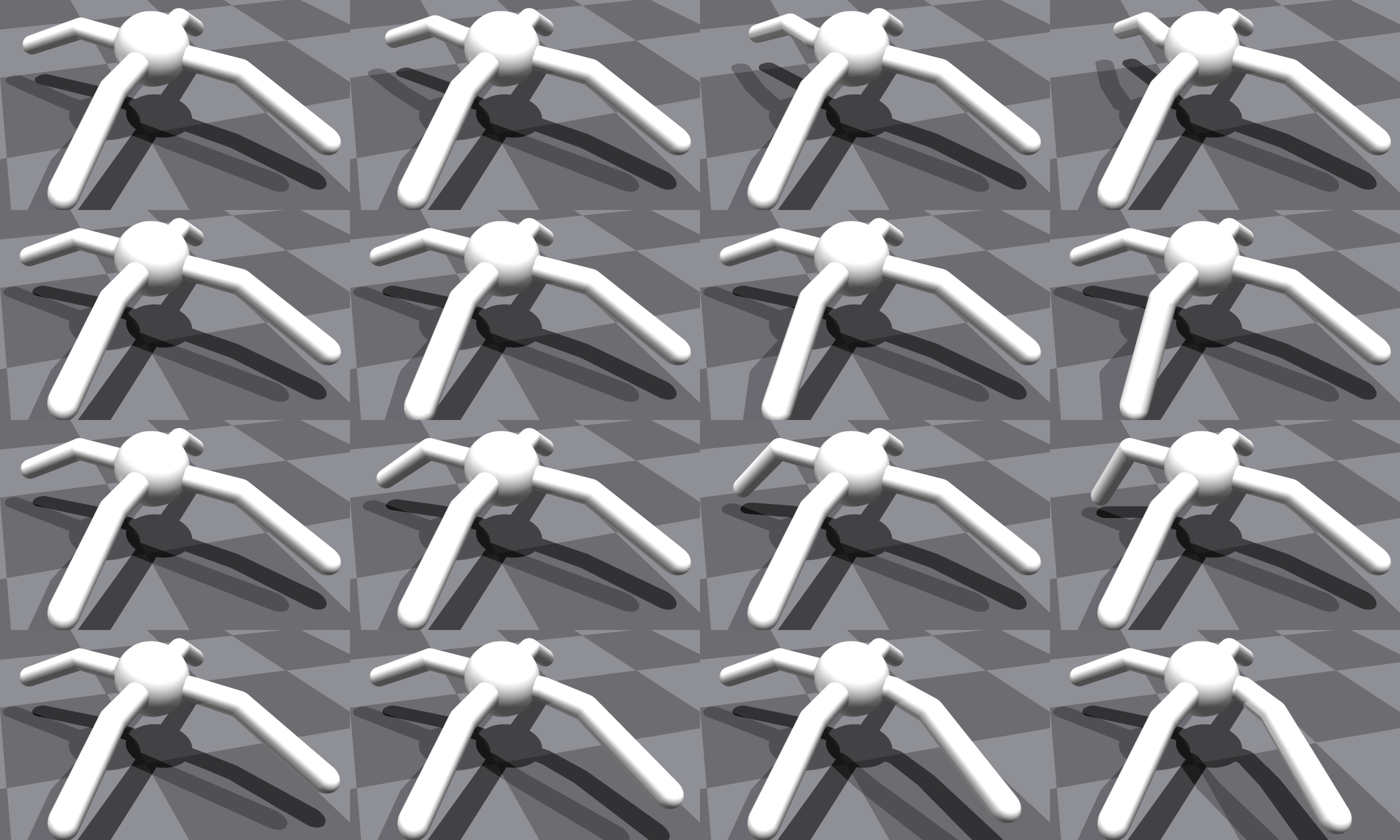}
        \subcaption{DOFs 4-7.}
    \end{minipage}%
    \caption{DOFs of Ant.}
\label{fig:dofs_ant}
\end{figure}

\begin{figure*}[ht]
\centering 
\includegraphics[width=\linewidth, trim=0 0 0 0, clip]{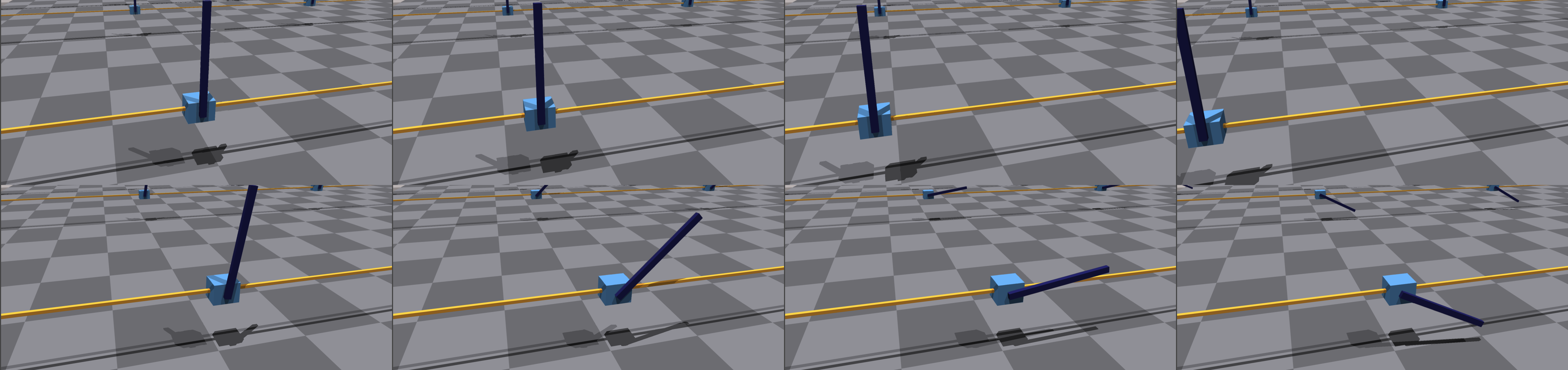}
\caption{DOFs of CartPole.}
\label{fig:dofs_cartpole}
\end{figure*}

\begin{figure}[ht]
    \centering
    \begin{minipage}{.49\linewidth}
        \includegraphics[width=\linewidth, trim=0 0 0 0, clip]{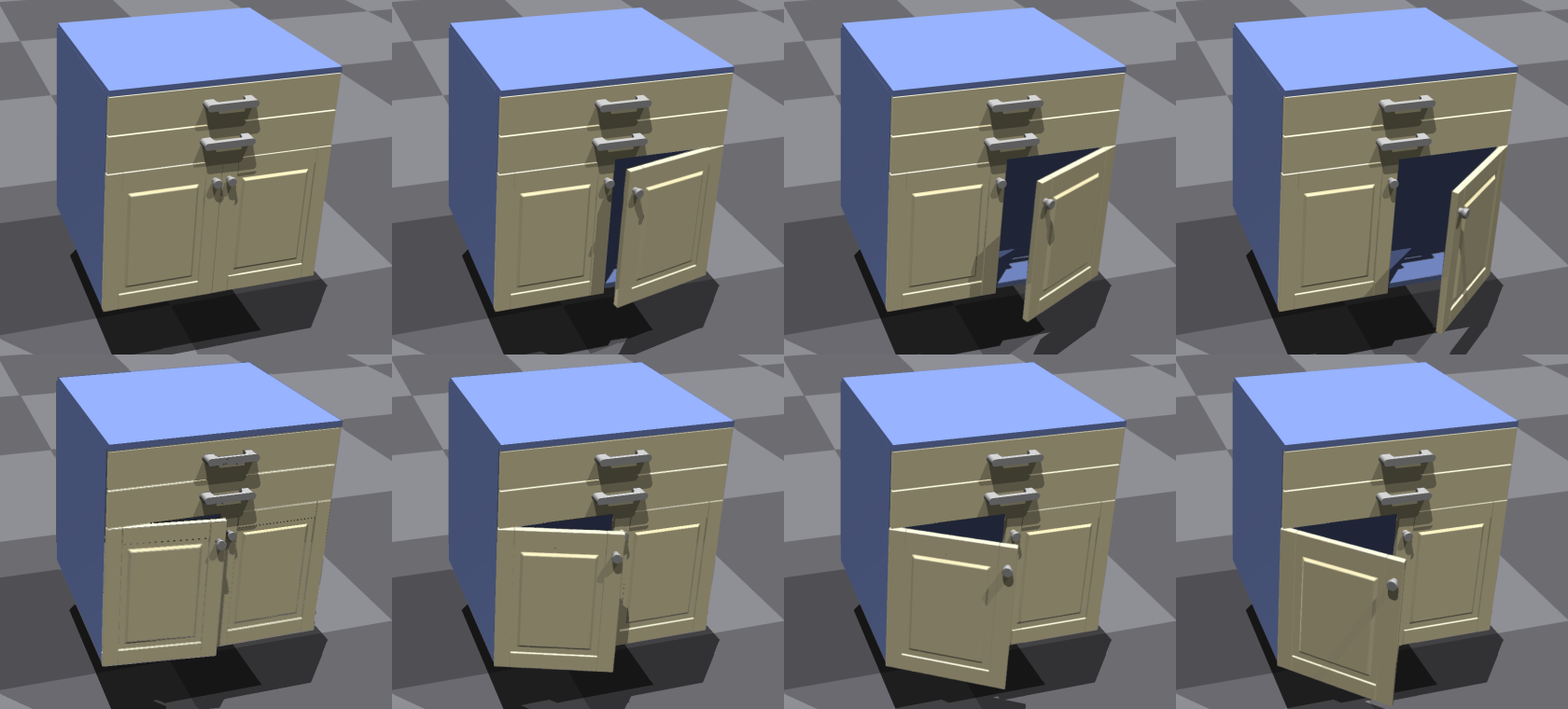}
        \subcaption{DOFs 0-1.}
    \end{minipage}%
    \begin{minipage}{.49\linewidth}
        \includegraphics[width=\linewidth, trim=0 0 0 0, clip]{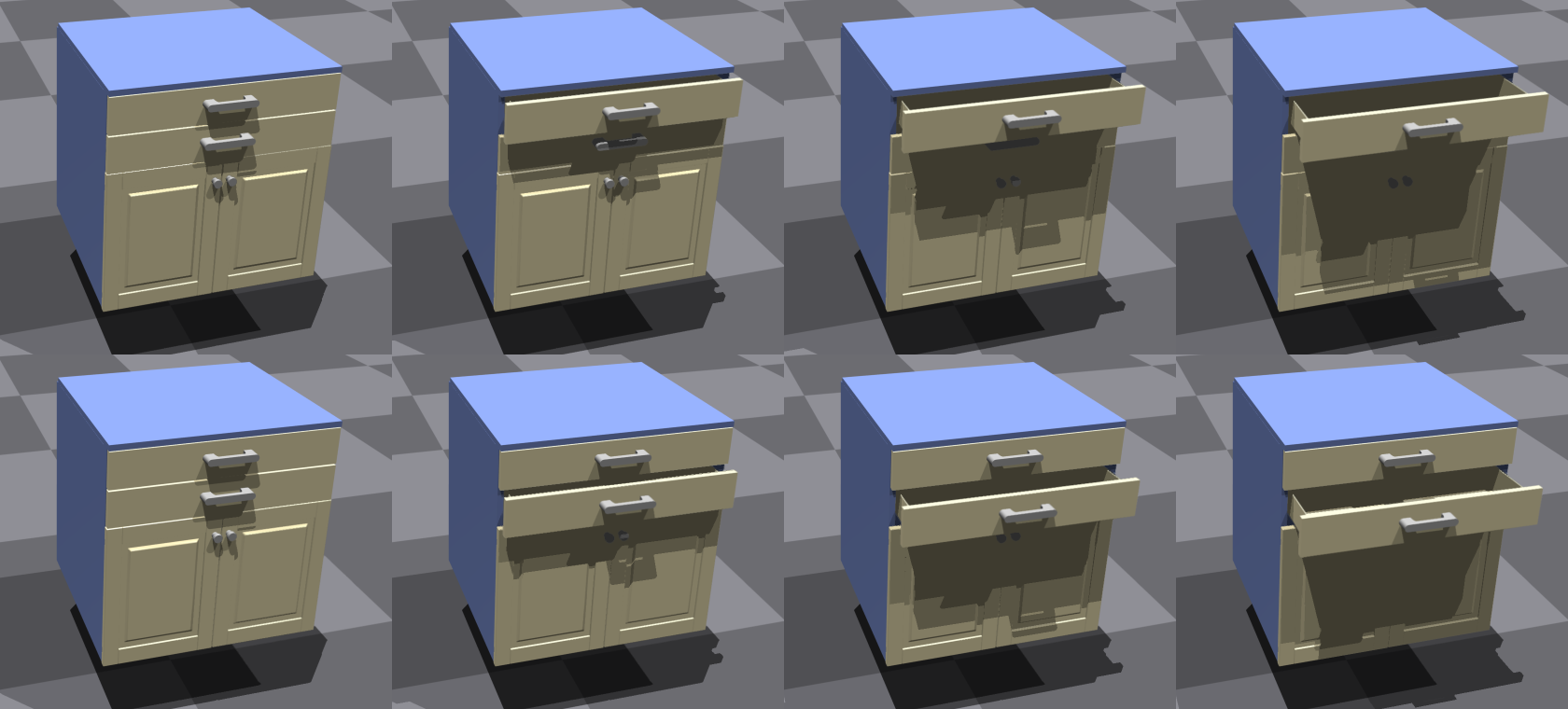}
        \subcaption{DOFs 2-3.}
    \end{minipage}%
\caption{DOFs of Sektion Cabinet.}
\label{fig:dofs_cabinet}
\end{figure}

\begin{figure}[ht]
    \centering
    \begin{minipage}[t]{.4\linewidth}
        \includegraphics[width=\linewidth, trim=0 0 0 0, clip]{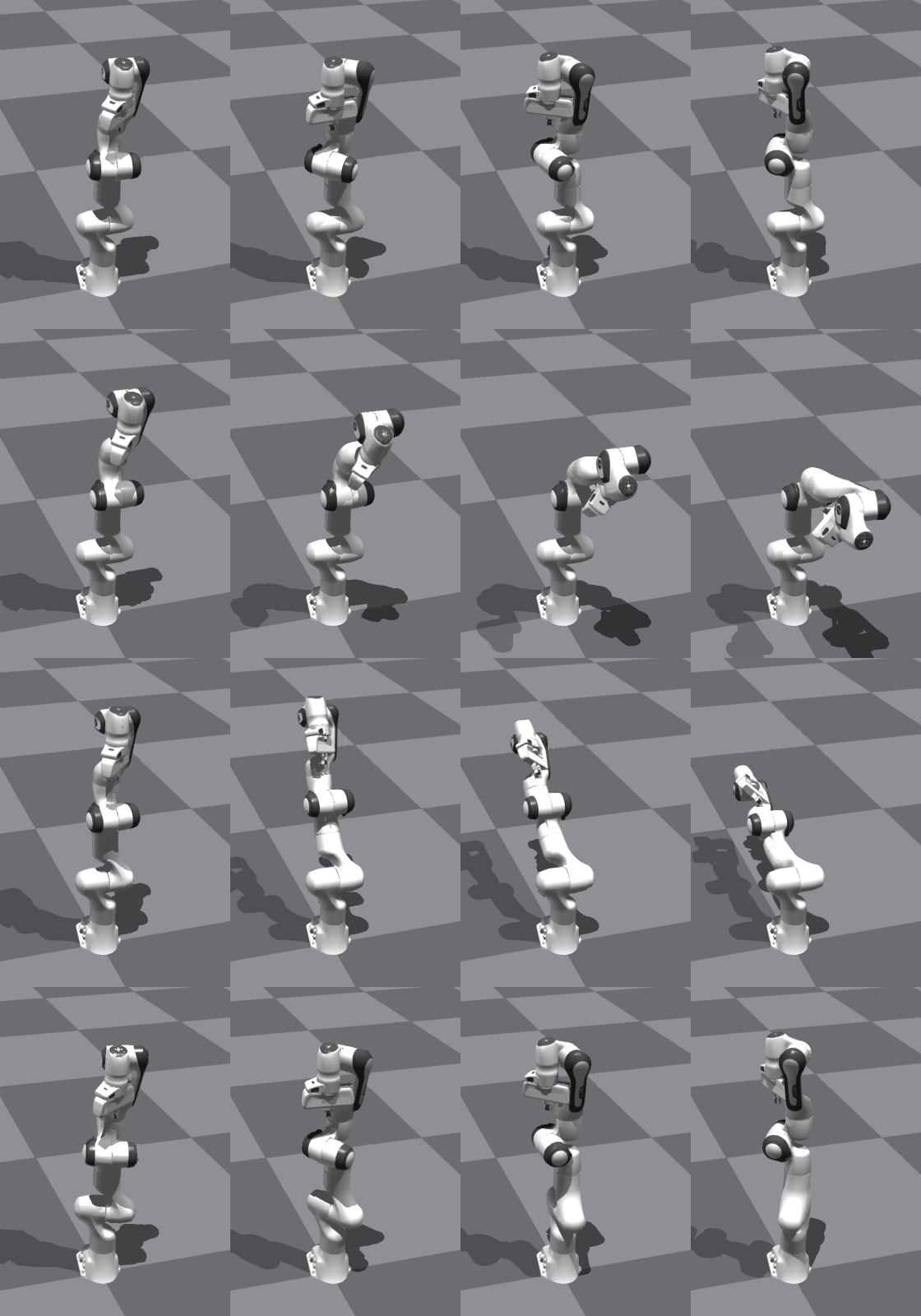}
        \subcaption{DOFs 0-3.}
    \end{minipage}%
    \begin{minipage}[t]{.4\linewidth}
        \includegraphics[width=\linewidth, trim=0 0 0 0, clip]{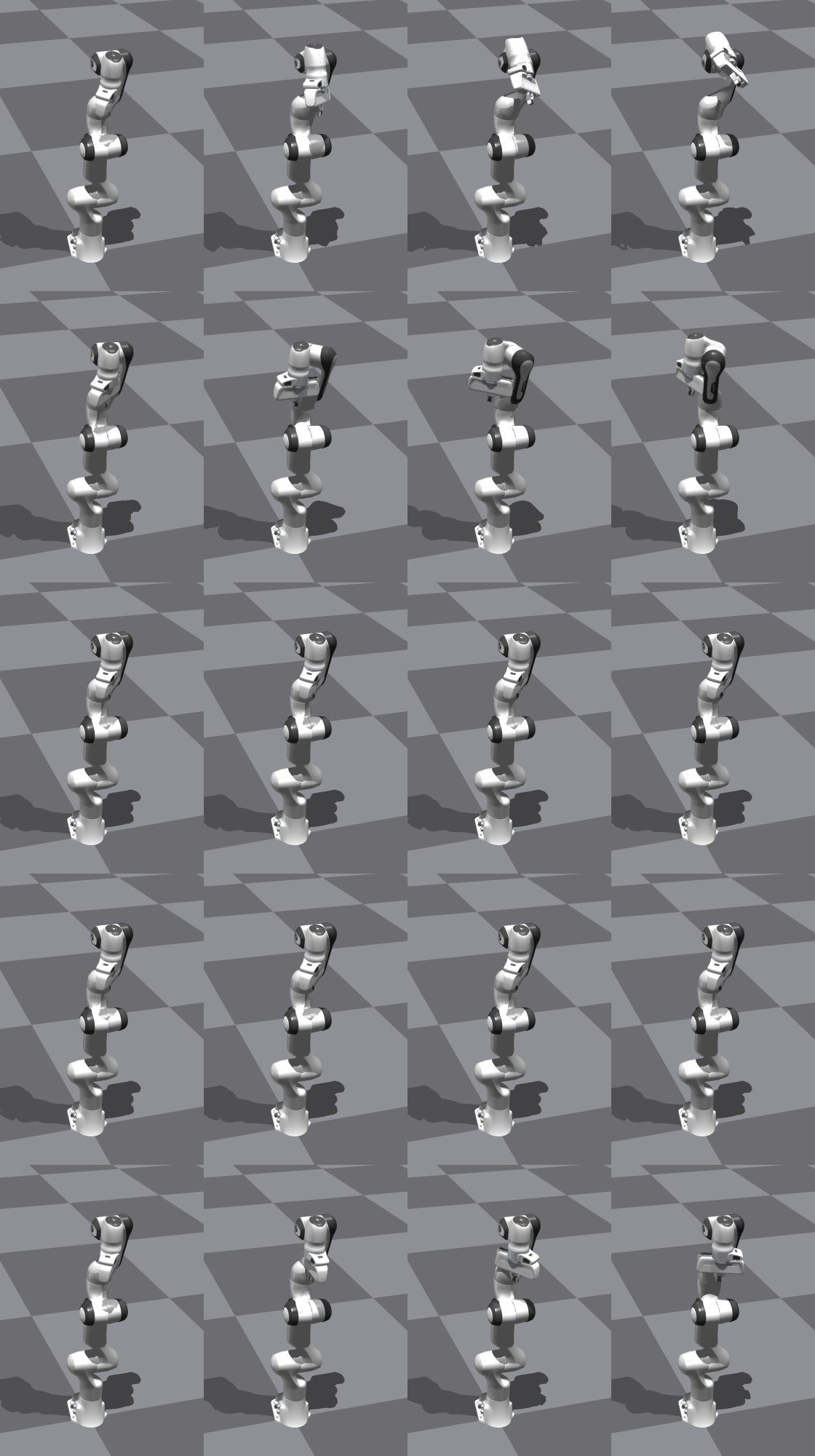}
        \subcaption{DOFs 4-8.}
    \end{minipage}%
\caption{DOFs of Franka Panda.}
\label{fig:dofs_panda}
\end{figure}

\begin{figure}[ht]
    \centering
    \begin{minipage}[t]{.4\linewidth}
        \includegraphics[width=\linewidth, trim=0 0 0 0, clip]{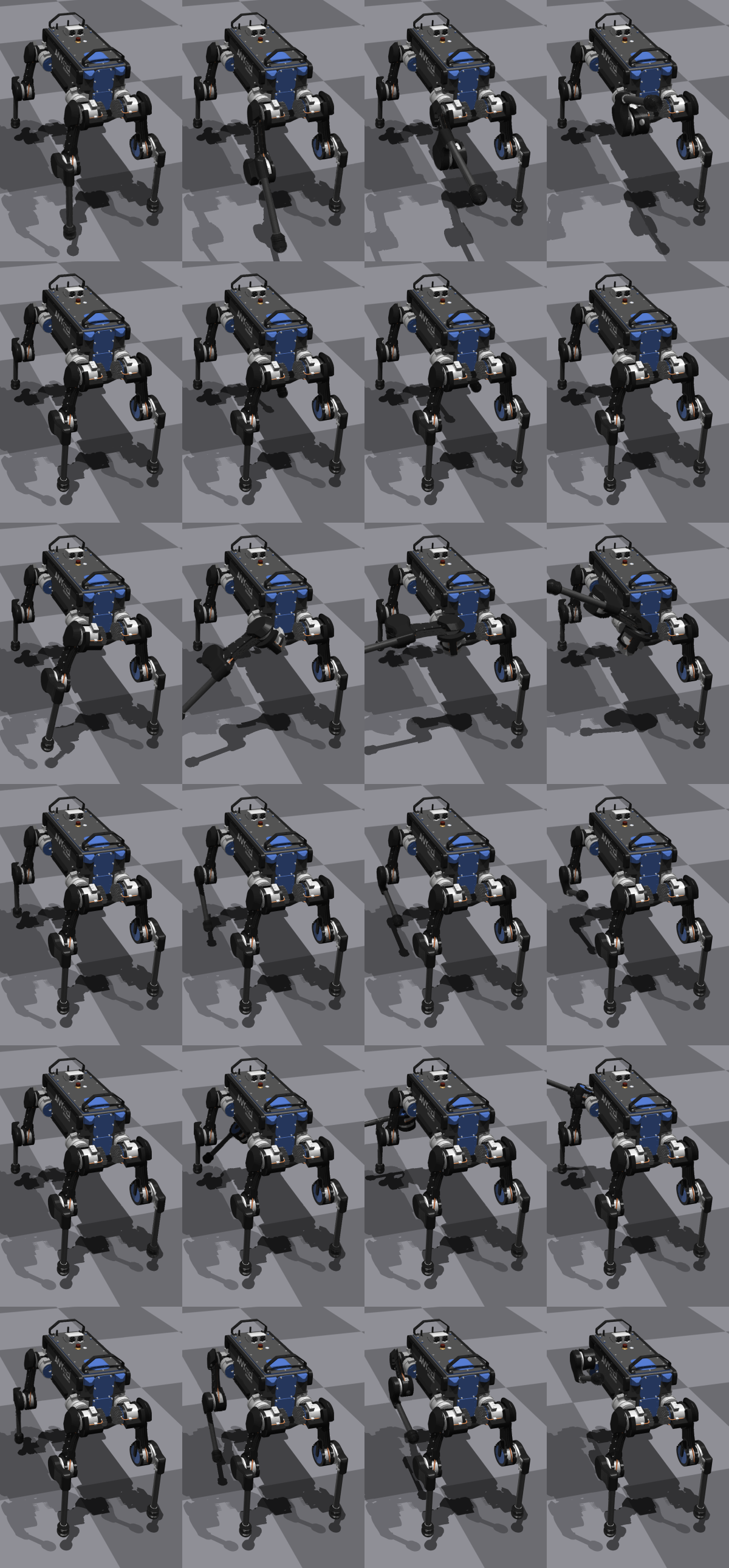}
        \subcaption{DOFs 0-5.}
    \end{minipage}%
    \begin{minipage}[t]{.4\linewidth}
        \includegraphics[width=\linewidth, trim=0 0 0 0, clip]{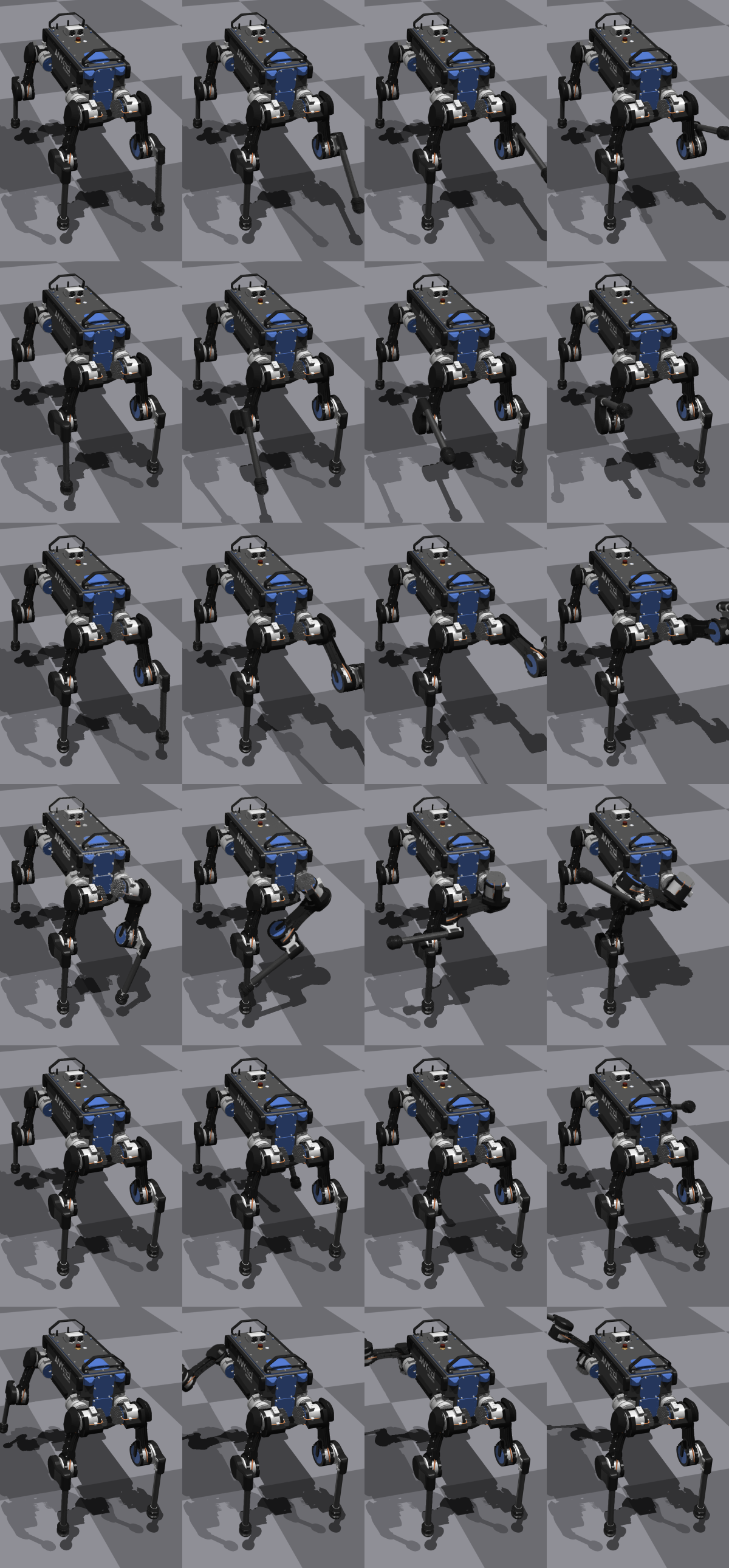}
        \subcaption{DOFs 6-11.}
    \end{minipage}%
\caption{DOFs of Anymal.}
\label{fig:dofs_anymal}
\end{figure}

%%%%%%%%%%%%%%%%%%%%%%%%%%%%%%%%%%%%%%%%%%%%%%%%%%%%%%%%%%%%%%%%
In this section, we present a series of diagrams including the Human~(\cref{fig:dofs_human}), Ant~(\cref{fig:dofs_ant}), Cartpole~(\cref{fig:dofs_cartpole}), Sektion Cabinet~(\cref{fig:dofs_cabinet}) Franka Panda~(\cref{fig:dofs_panda}), Kinova Cabinet~(\cref{fig:dofs_cabinet}), and Anymal~(\cref{fig:dofs_anymal}). The figures visually provides a clear graphical representation of its Degrees of Freedom (DOFs). Through these figures, readers enhance their comprehension of various DOFs.

\begin{figure}[ht]
    \centering
    \begin{minipage}[t]{.4\linewidth}
        \includegraphics[width=\linewidth, trim=0 0 0 0, clip]{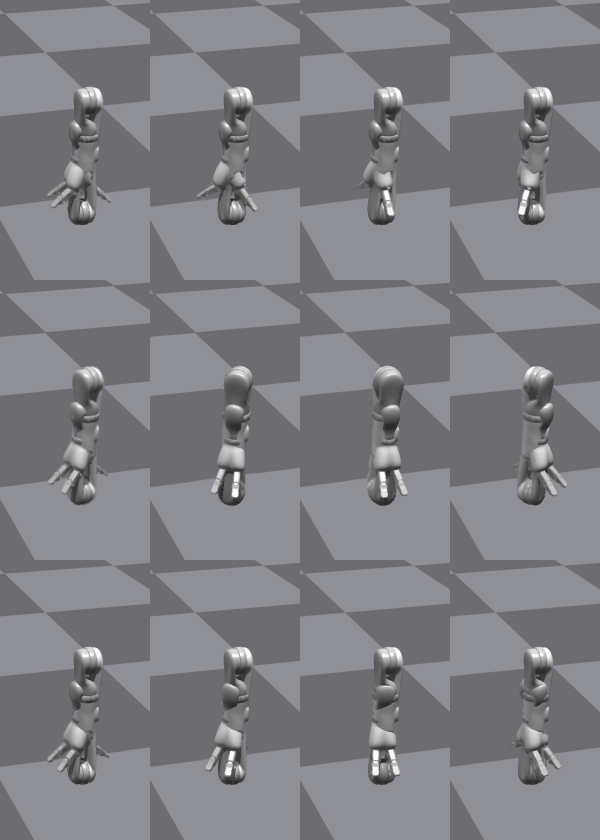}
        \subcaption{DOFs 0-2.}
    \end{minipage}%
    \begin{minipage}[t]{.4\linewidth}
        \includegraphics[width=\linewidth, trim=0 0 0 0, clip]{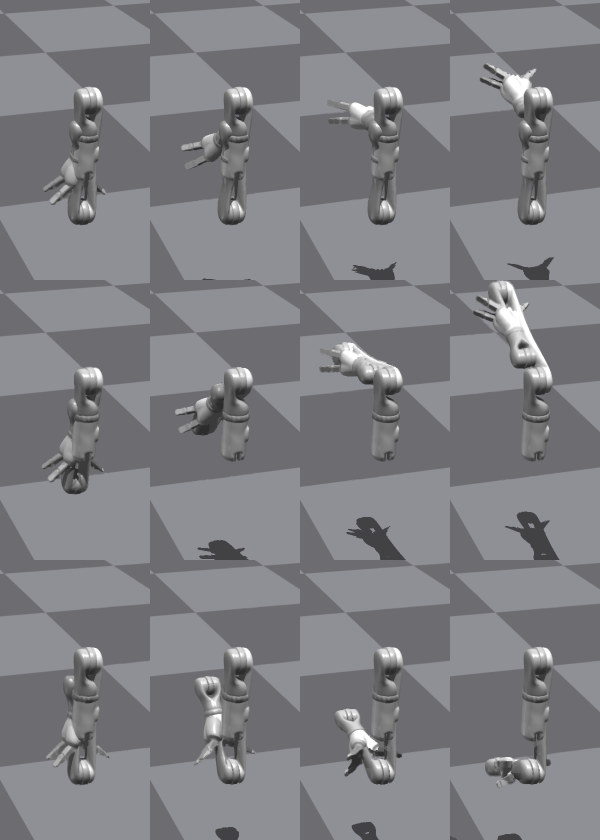}
        \subcaption{DOFs 3-5.}
    \end{minipage}%
\caption{DOFs of Kinova.}
\label{fig:dofs_kinova}
\end{figure}

\clearpage
%%%%%%%%%%%%%%%%%%%%%%%%%%%%%%%%%%%%%%%%%%%%%%%%%%%%%%%%%%%%%%%%
\section{Test Space}
\label{tab:test_space}
%%%%%%%%%%%%%%%%%%%%%%%%%%%%%%%%%%%%%%%%%%%%%%%%%%%%%%%%%%%%%%%%

In this section, we introduce a testing space to evaluate the performance of our newly proposed baseline framework across diverse entities. These entities, namely \textit{Human, Ant, Cartpole, Sektion Cabinet, Franka Panda, Kinova}, and \textit{Anymal} have been selected for the unique characteristics and challenges they present, making them ideal candidates for a comprehensive assessment.

For example, one of the tables showcases a variety of tasks for \textit{Human} entity, ranging from basic movements such as raising an arm to more complex activities like running. These tasks are intended to probe the baseline framework's ability to simulate and understand the complexities of human motion accurately.

The primary purpose of these tables is twofold: firstly, to offer a clear and structured means of testing our baseline framework's capability in replicating and predicting the dynamics of various movements and interactions within and across different entities. Secondly, they are intended to serve as a benchmark for future research endeavors. By establishing a set of standardized tasks, we aim to encourage and facilitate subsequent studies to build upon and innovate beyond our baseline framework, pushing the boundaries of what is currently possible in the field.

\begin{longtable}{lp{10cm}}
\caption{Tasks by Entity}\\
\toprule
\textbf{Entity} & \textbf{Tasks} \\
\midrule
\endfirsthead

\multicolumn{2}{c}%
{\tablename\ \thetable\ -- \textit{Continued from previous page}} \\
\toprule
\textbf{Entity} & \textbf{Test Tasks} \\
\midrule
\endhead

\multicolumn{2}{r}{\textit{Continued on next page}} \\
\endfoot

\bottomrule
\endlastfoot

Human & \textit{ raise your left arm, raise both your arms, put your left hand on left hip, place your left feet on right feet, place your right feet on left feet, put your left hand on left shoulder, put your left hand on left waist, put your left hand on right knee, rotate the right elbow, rotate the right knee, rotate the left ankle, rotate the left hip, rotate the right hip, stretch right arm forward, stretch left arm forward, swing the right arm, swing the left arm, run }\\
\midrule
Ant & \textit{ clockwise rotate the left bottom hip, clockwise rotate the right bottom hip, clockwise rotate the left upper hip, clockwise rotate the right upper hip, anticlockwise rotate the left bottom hip, anticlockwise rotate the right bottom hip, anticlockwise rotate the left upper hip, anticlockwise rotate the right upper hip }\\
\midrule
Cartpole & \textit{ left move the slider, right move the slider, anticlockwise rotate the cart, left and then right move the slider, right and then left move the slider, anticlockwise and then clockwise rotate the cart, clockwise and then anticlockwise rotate the cart }\\
\midrule
Sektion Cabinet & \textit{ open the left door, open the right door, open the bottom drawer, open and then close the left door, open and then close the right door, open and then close the bottom drawer, open and then close the front drawer }\\
\midrule
Franka Panda & \textit{ clockwise rotate the first joint, anticlockwise rotate the first joint, clockwise rotate the second joint, anticlockwise rotate the second joint, clockwise rotate the third joint, anticlockwise rotate the third joint, clockwise rotate the forth joint, clockwise rotate the fifth joint, anticlockwise rotate the fifth joint, clockwise rotate the sixth joint, anticlockwise rotate the sixth joint, clockwise rotate the seventh joint, anticlockwise rotate the seventh joint, bend the first finger, extend the first finger, bend the second finger, extend the second finger }\\
\midrule
Kinova & \textit{ anticlockwise rotate the first joint, clockwise rotate the first joint, anticlockwise swing the second joint, clockwise swing the second joint, anticlockwise swing the third joint, clockwise swing the third joint, anticlockwise rotate the forth joint, clockwise rotate the forth joint, anticlockwise swing the fifth joint, clockwise swing the fifth joint, anticlockwise rotate the sixth joint, clockwise rotate the sixth joint }\\
\midrule
Anymal & \textit{ walk, forward swing right front hip, backward swing right front hip, forward swing right front hip, backward swing left front hip, forward swing right hind hip, backward swing right hind hip, forward swing right hind hip, right swing the right hind hip, right swing the right front hip, left swing the right front hip }\\
\end{longtable}
\label{tab:test_space}

%\clearpage
%%%%%%%%%%%%%%%%%%%%%%%%%%%%%%%%%%%%%%%%%%%%%%%%%%%%%%%%%%%%%%%%
\section{Action Space}
\label{sec:action_space}
%%%%%%%%%%%%%%%%%%%%%%%%%%%%%%%%%%%%%%%%%%%%%%%%%%%%%%%%%%%%%%%%

\begin{figure*}[ht]
\centering 
\includegraphics[width=\textwidth, trim=0 0 0 0, clip]{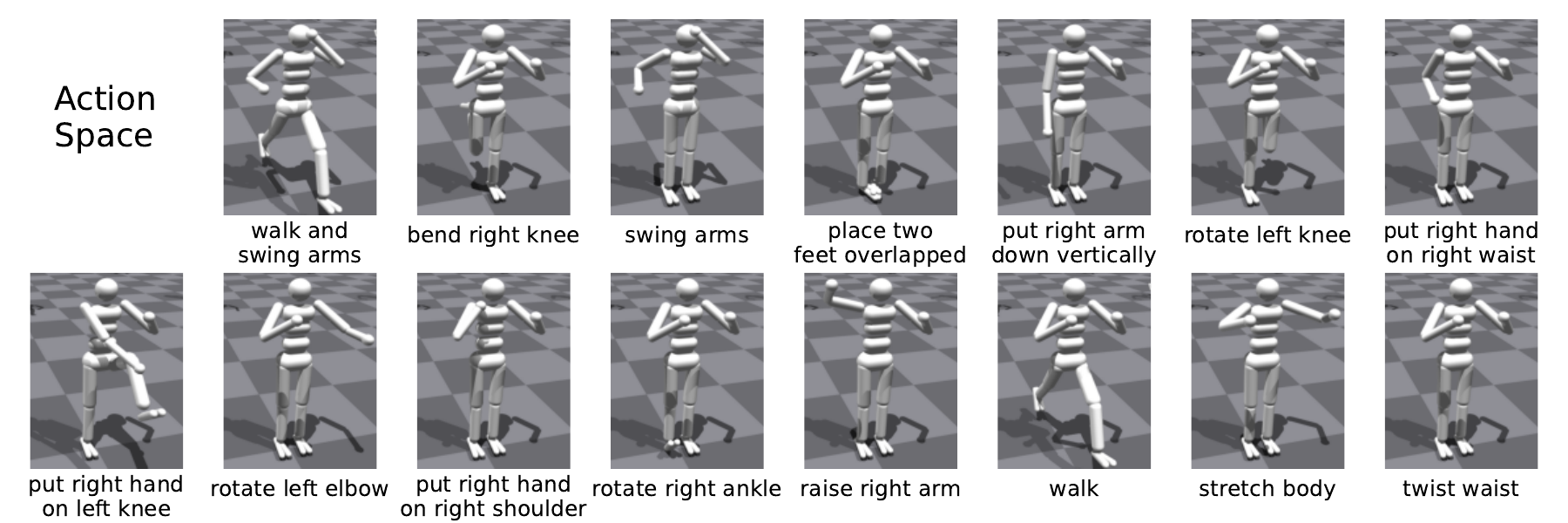}
\caption{The illustration of the action space of Human.}
\label{fig:action_space}
\end{figure*}

\cref{tab:action_space} plays a pivotal role as the initial action space for our framework, which is ingeniously designed to enhance the model's capabilities in an open-world setting. Each row within the table meticulously details the various actions associated with entities across different entities, including Human, Ant, Cartpole, Sektion Cabinet, Franka Panda, Kinova, and Anymal. These actions range from simple movements such as "stretch the body" and "walk" for the human entity, to more specific actions like "open top drawer" for the Sektion Cabinet, illustrating the diverse range of interactions that our model aims to understand and simulate. For example, we illustate the action space of human in \cref{fig:action_space}.

Our framework's objective is to dynamically update and improve this action space, thereby significantly enriching the model's adaptability and performance in complex, ever-changing environments. By starting with this foundational action space, we provide a benchmark for the initial capabilities of our model. However, through continuous learning and interaction with the open world, the framework is expected to evolve, expanding its repertoire of actions to accommodate new challenges and scenarios it encounters.

\begin{table}[h]
\centering
\caption{Actions by Entities}
\setlength{\tabcolsep}{5mm}
\begin{tabular}{ll}
\toprule
Human & \textit{ stretch the body }\\
& \textit{ put your right hand on right waist }\\
& \textit{ walk }\\
& \textit{ rotate the left elbow }\\
& \textit{ put your right hand on right shoulder }\\
& \textit{ put your right arm down vertically }\\
& \textit{ walk and swing arms }\\
& \textit{ raise your right arm }\\
& \textit{ rotate the right ankle }\\
& \textit{ place the two feet overlapped }\\
& \textit{ rotate the left knee }\\
& \textit{ bend the right knee }\\
& \textit{ swing arms }\\
& \textit{ put your right hand on the left knee }\\
& \textit{ twist the waist }\\
\midrule
Ant & \textit{ bend the right upper knee }\\
& \textit{ bend the right bottom knee }\\
\midrule
Cartpole & \textit{ clockwise rotate pole }\\
\midrule
Sektion Cabinet & \textit{ open top drawer }\\
\midrule
Franka Panda & \textit{ clockwise rotate 1 dof }\\
\midrule
Kinova & \textit{ clockwise rotate 1 dof }\\
\midrule
Anymal & \textit{ extend the left front knee }\\
& \textit{ bend left front knee }\\
& \textit{ left swing the right hind hip }\\
\bottomrule
\end{tabular}
\label{tab:action_space}
\end{table}

%%%%%%%%%%%%%%%%%%%%%%%%%%%%%%%%%%%%%%%%%%%%%%%%%%%%%%%%%%%%%%%%
\section{Pass Rates of Code Evaluator}
\label{sec:code_evaluator}
%%%%%%%%%%%%%%%%%%%%%%%%%%%%%%%%%%%%%%%%%%%%%%%%%%%%%%%%%%%%%%%%

\begin{figure*}[ht]
\centering 
\includegraphics[width=\textwidth, trim=0 0 0 0, clip]{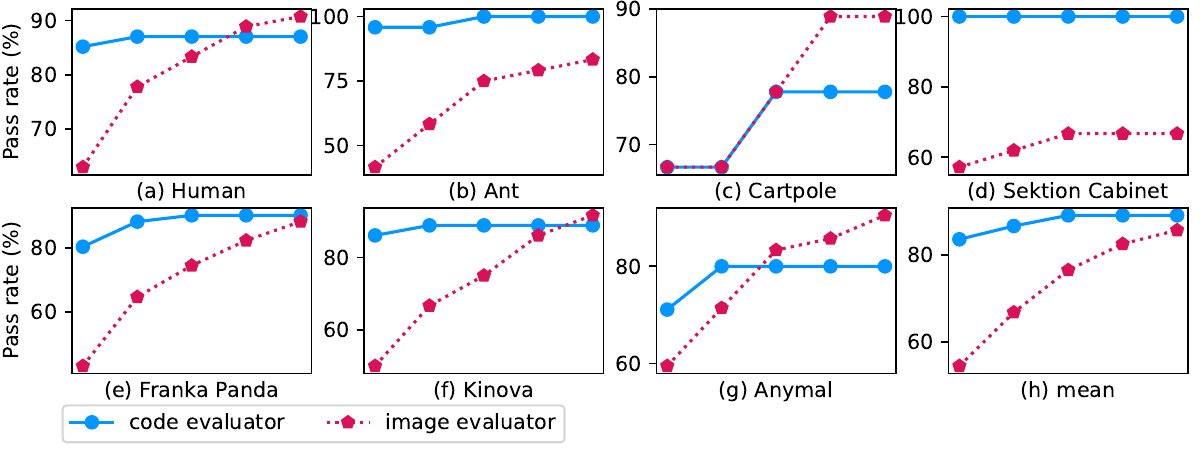}
\caption{The pass rate of Image Evaluator and Code Evaluator.}
\label{fig:code_evaluator}
\end{figure*}

In \cref{fig:code_evaluator}, we compare the pass rates as determined by the Image Evaluator and the Code Evaluator. The findings indicate that the Code Evaluator exhibits a high pass rate within the first one or two iterations, suggesting that it tends to mistakenly believe the code has already accomplished the task. Consequently, employing the Code Evaluator as a criterion for concluding the loop and providing feedback would result in the iteration process being prematurely terminated at its initial stages, significantly limiting the potential for model performance enhancement. This experiment underscores the superiority of our proposed Image Evaluator.

\section{Update Action Space}
A strategy for updating the action space is serial updating (\cref{tab:serial}), which involves updating the action space after each task is completed or when the maximum number of iterations is reached, before proceeding to the next task. In contrast, we employ a parallel updating strategy (\cref{tab:parallel}), where all tasks are tested simultaneously. Simple tasks can be quickly integrated into the action space, while difficult tasks can benefit from insights gained from simpler ones. After each iteration, we update the action space before moving forward.
\begin{figure}[t]
\centering
\begin{subtable}{.5\linewidth}
\centering
\begin{tabular}{c|c|ccccc|ccc|c}
\toprule
Iteration & 1 & 1 & 2 & 3 & 4 & 5 & 1 \\
\midrule
Task 1 & \ding{51} & & & & & & \\
Task 2 & & \ding{55} & \ding{55} & \ding{55} & \ding{55} & \ding{55} \\
Task 3 & & & & & & & \ding{55}  \\
\bottomrule
\end{tabular}
\caption{Serial Update}
\label{tab:serial}
\end{subtable}%
\begin{subtable}{.5\linewidth}
\centering
\begin{tabular}{c|ccccc|ccccc}
\toprule
Iteration & 1 & 2 & 3 & 4 & 5 & 1 \\
\midrule
Task 1 & \ding{51} & & & & & \\
Task 2 & \ding{55} & \ding{55} & \ding{55} & \ding{55} & \ding{55} & \ding{51} \\
Task 3 & \ding{55} & \ding{55} & \ding{51} & & & \\
\bottomrule
\end{tabular}
\caption{Parallel Update}
\label{tab:parallel}
\end{subtable}
\captionof{table}{Examples of Serial and Parallel Update.}
\end{figure}

\end{document}